\documentclass[journal]{IEEEtran}

\usepackage{graphicx}
\usepackage{amsmath}
\usepackage{amssymb}
\usepackage{xcolor}
\usepackage{verbatim}
\usepackage{hyperref}
\usepackage{algorithm}
\usepackage{algpseudocode}
\usepackage{listings}
\usepackage{caption}
\usepackage{subcaption}
\usepackage{authblk}
\usepackage{multirow}
\usepackage{amsthm}
\usepackage{amsmath}
\usepackage{makecell}
\usepackage{yfonts}
\usepackage[noadjust]{cite}
\usepackage[labelformat=simple]{subcaption}

\DeclareMathOperator*{\argminA}{arg\,min}

\def\x{\boldsymbol{x}}
\def\m{\mathfrak{m}}
\def\real{\mathbb{R}}

\def\S{\mathcal{S}}
\def\A{\mathcal{A}}

\def\v{\mathbf{v}}

\def\p{\mathbf{p}}
\def\a{\mathbf{a}}
\def\tf{\tilde{f}}

\newtheorem{lemma}{Lemma}
\newtheorem{definition}{Definition}

\newtheorem{prop}{Proposition}
\newtheorem{assumption}{Assumption}

\begin{document}

\title{Manifoldron: Direct Space Partition via Manifold Discovery}

\author{
Dayang Wang$^{1\dag}$,
Feng-Lei Fan$^{2\dag}$, \textit{IEEE Member},
Bo-Jian Hou$^2$,
Hao Zhang$^2$, 
Zhen Jia$^3$,
Boce Zhang$^3$,\\
Rongjie Lai$^{4*}$,
Hengyong Yu$^{1*}$, \textit{IEEE Senior Member},
Fei Wang$^{2*}$, \textit{IEEE Senior Member}\thanks{$^*$Rongjie Lai, Hengyong Yu, and Fei Wang are co-corresponding authors. $^\dag$Dayang Wang and Feng-Lei Fan contribute equally.}
\thanks{
$^1$Dayang Wang and Hengyong Yu are with Department of Electrical and Computer Engineering, University of Massachusetts, Lowell, MA, USA}
\thanks{$^2$Feng-Lei Fan, Bo-Jian Hou, Hao Zhang, and Fei Wang are with Weill Cornell Medicine, Cornell University, New York City, NY, USA}
\thanks{$^3$Zhen Jia and Boce Zhang are with Department of Food Science and Human Nutrition, University of Florida, Gainesville, FL, USA}
\thanks{$^4$Rongjie Lai is with Department of Mathematics, Rensselaer Polytechnic Institute, Troy, NY, USA}}


\maketitle

\begin{abstract}
A neural network with the widely-used ReLU activation has been shown to partition the sample space into many convex polytopes for prediction. However, the parameterized way a neural network and other machine learning models use to partition the space has imperfections, \textit{e}.\textit{g}., the compromised interpretability for complex models, the inflexibility in decision boundary construction due to the generic character of the model, and the risk of being trapped into shortcut solutions. In contrast, although the non-parameterized models can adorably avoid or downplay these issues, they are usually insufficiently powerful either due to over-simplification or the failure to accommodate the manifold structures of data. In this context, we first propose a new type of machine learning models referred to as Manifoldron that directly derives decision boundaries from data and partitions the space via manifold structure discovery. Then, we systematically analyze the key characteristics of the Manifoldron such as manifold characterization capability and its link to neural networks. The experimental results on 4 synthetic examples, 20 public benchmark datasets, and 1 real-world application demonstrate that the proposed Manifoldron performs competitively compared to the mainstream machine learning models. We have shared our code in \url{https://github.com/wdayang/Manifoldron} for free download and evaluation.
\end{abstract}

%
\IEEEpeerreviewmaketitle

\section{Introduction}

Over the past few years, deep learning has transformed a huge variety of important applications \cite{lecun2015deep,fan2018learning,lin2021fully,wang2021ted}. With the development of research, the understanding of the inner-working mechanism of a deep neural network is gradually enhanced from both theoretical and engineering perspectives. One notable thread of studies \cite{chu2018exact,balestriero2020mad,fan2020quasi} mathematically showed that a neural network with the widely-used ReLU activation partitions the sample space into many convex polytopes, and a neural network is essentially a piecewise linear function over these polytopes. In theory, a polytope and its underpinning linear function can be uniquely determined by the collective activation states of each and every neuron in a network \cite{chu2018exact}, thereby realizing a complete and consistent interpretation of the original network. However, in practice, it is unrealistic to do so because the computational cost scales exponentially with the number of neurons. 

Space partition is not a proprietary view exclusively owned by deep learning; it is also shared by other machine learning models. We categorize machine learning models as parameterized and non-parameterized ones, which respectively correspond to encoding decision boundaries with model parameters to partition the space (parameterized) and directly partitioning the space with training data (non-parameterized). 

Compared to direct space partition, the use of parameterized decision boundaries is subjected to three limitations. (i) The compromised interpretability \cite{fan2021interpretability,hou2020learning}: As the model goes complex, \textit{e}.\textit{g}., a neural network gets deeper, or a decision tree \cite{myles2004introduction} is ensembled into a forest, how model parameters shape the decision boundaries becomes more and more unclear, making the entire model less and less interpretable. (ii) The parameterized boundaries have to respect a certain standard about the form of the decision boundaries that concerns the generic character of the model, \textit{e}.\textit{g}., a single decision tree typically employs horizontal and vertical decision hyperplanes. These standards may not facilitate learning complicated patterns. (iii) Many studies \cite{bai2021attentions,robinson2021can} have observed that the use of parameterized decision boundaries can be trapped into shortcut solutions instead of mining genuine features, because a shortcut solution resides in a large basin in the landscape of the loss function. For example, let us utilize a neural network to recognize 0-1 digits. If we purposely shift all 0-digits to the left region of an image and 1-digits to the right, the training of a neural network could probably learn parameters to make decisions just based on the location rather than digit features.


\begin{figure}[tb]
\centering
\includegraphics[width=\linewidth]{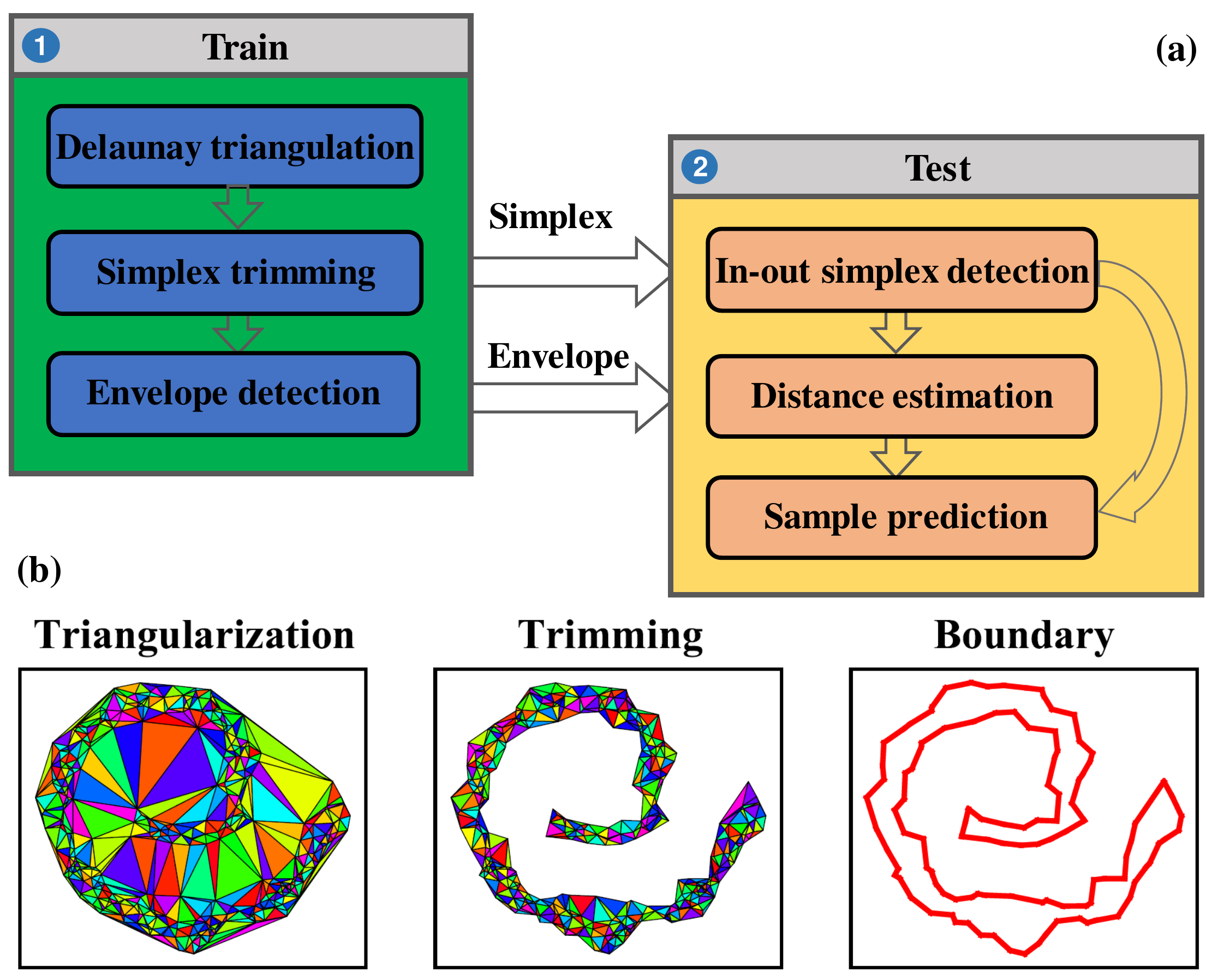}
\caption{(a) The pipeline of the proposed Manifoldron. In the training stage, Delaunay triangulation, simplex trimming, and envelope detection are sequentially performed to identify the manifold structure of data and establish decision boundaries. In the test stage, we classify new points based on in-out simplex detection and distance estimation. (b) An exemplary illustration of the Manifoldron.} \label{fig:pipeline_and_example}
\vspace{-0.4cm}
\end{figure}

In contrast, non-parameterized models directly partition the space with the training data. For example, the k-nearest neighbors algorithm \cite{biau2015lectures} classifies a sample according to how its k-nearest neighbors are classified, and the Naive Bayes model \cite{murphy2006naive} classifies a point with Bayes theorem on conditional distribution between the class and features. Although these models can avoid or downplay the issues of parameterized decision boundaries, they are usually insufficiently powerful either due to over-simplification or the failure to accommodate the manifold structures of data. Our motivation is that can we develop a model that not only partitions the space in a non-parameterized way but also is sufficiently capable? 

To answer this question, in this paper, we first propose a new type of classifiers referred to as Manifoldron that directly derives decision boundaries to partition the space from data via manifold discovery. Specifically, as Figure \ref{fig:pipeline_and_example} shows, we triangulate the data of each class and trim the resultant triangulation to gain a compact envelope of the manifold of data. Naturally, the envelopes of all classes will serve as the decision boundaries for classification. We also generalize the Manifoldron from a classifier to a regressor by presenting a novel and concise regression formula. Furthermore, we propose to use feature bagging and point2point projection aided by parallel computation to greatly speed up the training and inference of the Manifoldron. Next, as a necessary step to roll out a new model, we systematically analyze the key characteristics of the Manifoldron, \textit{e}.\textit{g}., manifold characterization capability and its link to neural networks. These characteristics also elaborate how and why the Manifoldron can overcome the issues of parameterized decision boundaries. Lastly, we apply the Manifoldron to 4 synthetic examples, 20 benchmark datasets, and 1 real-world application to validate the advantages of the model and the model’s key components. 

\textbf{Main contributions.} (i) We propose a novel type of machine learning models called Manifoldron for classification tasks. (ii) We demonstrate the key properties of the Manifoldron such as manifold characterization capability and the link to neural networks to pave the way for future applications. (iii) We conduct comprehensive experiments to validate that the Manifoldron performs competitively compared to other mainstream machine learning models. 

\section{Related Work}
There are two lines of works that are highly relevant to ours.

\textbf{General machine learning models.} Despite different motivations, in the supervised setting, machine learning models essentially learn a function that maps an input to the desired output. The commonly-used models consist of the k-nearest neighbors (kNN), support vector machine (SVM) \cite{noble2006support}, Gaussian process (GP) \cite{rasmussen2002infinite}, decision tree, random forest (RF) \cite{biau2016random}, neural networks (NN), AdaBoost \cite{schapire2013explaining}, Naive Bayes, quadratic discriminant analysis (QDA) \cite{tharwat2016linear}, and so on. The Manifoldron is motivated by partitioning the space in a non-parameterized fashion, with an emphasis on leveraging the manifold structure of data. Our work confirms the feasibility of directly constructing decision boundaries by manifold discovery. To the best of our knowledge, the Manifoldron is not just another modification to the existing models but a fundamentally novel model for machine learning.

\textbf{Manifold learning} \cite{cayton2005algorithms} is a class of unsupervised and supervised models assuming that data lie in a low-dimensional manifold embedded in a high-dimensional space. Methodologically, the manifold assumption as a basic concept offers a ground-breaking view to many machine learning tasks such as dimension reduction \cite{roweis2000nonlinear,tenenbaum2000global}, representation learning \cite{bengio2013representation}, and so on. Practically, the manifold assumption can be embedded into the model to better characterize real-world problems. As a methodological innovation, our work is the first endeavor to directly use the manifold structure to construct a classifier, which has the potential to find interesting application niches in the future. 

\section{Manifoldron}

\subsection{Key Steps of the Manifoldron}

The key steps of the Manifoldron are shown in Figure \ref{fig:pipeline_and_example}. Now, let us introduce them in detail as follows. 


\begin{definition}[Simplicial complex, \cite{schonsheck2019chart}]
\label{def:simplex}
A $D$-simplex $\mathrm{S}$ is a $D$-dimensional convex hull provided by a convex combination of  $D+1$ affinely independent vectors $\{\v_d\}_{d=0}^D \subset \real^D$. In other words, $\displaystyle \mathrm{S} = \left\{ \sum_{d=0}^D \xi_d \v_d ~|~ \xi_d \geq 0, \sum_{d=0}^D \xi_d = 1 \right \}$.  The convex hull of any subset of $\{\v_d\}_{d=0}^D$ is called a \textit{face} of  $\mathrm{S}$. A simplicial complex $\displaystyle 
\mathcal{S} = \bigcup_\alpha \mathrm{S}_\alpha$ is composed of a set of simplices $\{S_\alpha\}$ satisfying: 1) every face of a simplex from $\S$ is also in $\S$; 2) the non-empty intersection of any two simplices $\displaystyle \mathrm{S}_{1},\mathrm{S}_{2}\in \S$ is a face of both $\mathrm{S}_1$ and $\mathrm{S}_2$. 
\end{definition}

\textbf{Delaunay triangulation.}
Given the point cloud of a certain class, a triangulation to it ends up with a simplicial structure $\bar{\mathcal{S}}$ whose profile is a convex hull. This simplicial structure and its associated convex hull completely cover the entire point cloud and the corresponding manifold (see Figure \ref{fig:pipeline_and_example}). Here, we employ the commonly-used Delaunay triangulation \cite{toth2017handbook} whose software implementation is conveniently accessible to the community. 

\textbf{Simplex trimming.}
As mentioned above, a Delaunay triangulation leads to a convex hull of the point cloud, which is usually not the true profile of the manifold. To resolve this issue, we trim simplices whose vertices are too far from one another. Mathematically, we build a KD tree \cite{ram2019revisiting}, which is a data structure facilitating neighborhood search, for all data beforehand. By presetting the number of neighbors and querying the KD tree to find neighbors, which has the time complexity of $\mathcal{O}(\log n)$, where $n$ is the number of samples, we delete those simplices in $\bar{\mathcal{S}}$ whose vertices are not neighbors, and the rest simplicial complex $\mathcal{S}$ is expected to reflect the manifold structure of data.

\textbf{Envelope detection.}
With $\mathcal{S}$ being prepared, we are ready to detect the envelope of the manifold. An envelope is a collection of all hyperplanes in the boundary. As we know, a boundary hyperplane is only shared by one simplex, while an interior hyperplane is shared by two simplices. Simply but effectively, we utilize this fact to determine if a hyperplane resides in a boundary and obtain the envelop $\mathcal{E}$.


\textbf{Sample prediction.}
For points of each class $c$, we have obtained the simplicial complex $\mathcal{S}_c$ representing the manifold and its envelope $\mathcal{E}_c$. Now we can utilize them for prediction. The prediction process is twofold: (i) in-out simplex detection, where we verify if a test sample lies within the envelope and (ii) distance estimation, where when a test sample is out of all envelopes, we find out which envelope $\mathcal{E}_c$ is closest to this test sample.

(i) \textit{In-out simplex detection.}
As mentioned earlier, a $D$-simplex $\mathrm{S}$ is a $D$-dimensional convex hull provided by a convex combination of  $D+1$ affinely independent vectors $\{\v_d\}_{d=0}^D \subset \real^D$. Mathematically, we have $\displaystyle \mathrm{S} = \left\{ \v =\sum_{d=0}^D \xi_d \v_d ~|~ \xi_d \geq 0, \sum_{d=0}^D \xi_d = 1 \right \}$, where $\xi_d,\ d=0,1,\cdots,D$ are the barycentric coordinates. Therefore, given a test point $\p$, we compute its barycentric coordinates by solving the following equations:
\begin{equation}
\begin{cases}
    & \sum_{d=0}^D \xi_d \v_d = \p\\
    & \sum^{D}_{d=1} \xi_d = 1. \\
\end{cases}
\label{eqn:in_out}
\end{equation}
If $\xi_d \geq 0, d=0,1,\cdots,D$, then the test point $\p$ lies in the simplex $\mathrm{S}$; otherwise, it does not. We repeat this detection process for all simplices.

(ii) \textit{Distance estimation}.
If a given point $\p$ is not lying within any envelope, we use the closest envelope $\mathcal{E}_c$ to $\p$ for classification. Mathematically, we define the distance between a point and an envelope as the minimum distance between the point and all hyperplanes pertaining to this envelope. Without loss of generality, the hyperplane $h$ of the envelope is denoted as
\begin{equation}
    \boldsymbol{w}^{(h)} \cdot \v^\top +b^{(h)} = 0,
    \label{eqn:distance}
\end{equation}
where $\boldsymbol{w}^{(h)}$ and $b^{(h)}$ are determined by vertices of $h$. Then, the distance between $\p$ and $h$ is calculated by the projection formula:
\begin{equation}
    \zeta^{(h)} = \frac{|\boldsymbol{w}^{(h)} \cdot \p^\top + b^{(h)}|}{\lVert \boldsymbol{w}^{(h)} \rVert},
\end{equation}
and the distance between $\p$ and $\mathcal{E}_c$ is defined as $\underset{h\in \mathcal{E}_c }{\min} \ \zeta^{(h)}$.  

In sample prediction, both the in-out simplex detection and distance estimation use the KD tree for neighborhood query such that we do not need to slowly numerate all simplices and hyperplanes of the envelope.

Typically, data manifolds are non-linearly separable
even if linearly non-separable. As long as we have sufficient data to identify the manifolds and associated envelopes, data points can be correctly classified, no matter if the manifolds are generated by a mixed model or not. Furthermore, we notice that the mixed model is quite general for data modeling, which fits the common datasets such as iris\footnote{\url{https://www.stats.ox.ac.uk/~sejdinov/teaching/dmml17/Mixtures.html}}, usps5\footnote{\url{https://en.wikipedia.org/wiki/Mixture_model}}, etc. Since our later experiments shows the effectiveness of the Manifoldron on iris, usps5, etc., we argue that our model can work on data generated from a mixture model.


\subsection{Computational Complexity Reduction}

\begin{table}[ht]
\caption{The computational saving of feature bagging and point2point projection, where $n$ is the number of samples, $D$ is the original dimension, and $D'$ is the dimension used in feature bagging.}
\centering

\begin{tabular}{||c|c||}
    \hline
    Strategy &   Major Effect \\ 
     \hline
     \hline
    \multirow{3}{*} {Feature Bagging} & Delaunay  Triangulation \\
   
      & $\mathcal{O}(n) \sim \mathcal{O}(n^{[D/2]})$ $\to$ \\ 
      
      & $\mathcal{O}(n) \sim \mathcal{O}(n^{[D'/2]})$ \\
      \cline{2-2}
     \hline
     \multirow{2}{*}{Point2Point Projection} & Projection\\
     
      & $\mathcal{O}(D'^2) \to \mathcal{O}(D')$  \\
    \hline
\end{tabular}
\label{tab:complexity}
\end{table}

Assume that $n$ is the number of samples, and $D$ is the feature dimension. The establishment of the Manifoldron might be computationally expensive if a data set has large $n$ and $D$. The bottlenecks are Delaunay triangulation and sample prediction. The former has the time complexity of $\mathcal{O}(n) \sim \mathcal{O}(n^{[D/2]})$ \cite{toth2017handbook}, where the best case is achieved for inputs chosen uniformly inside
a sphere, while the latter solves an equation slowly in deriving Eq. \eqref{eqn:distance}. To reduce the computational complexity and simultaneously retain the predictive accuracy, we propose two strategies. 

\textbf{Feature bagging.} Since the high dimensionality is a major booster of computational complexity, an intuitive idea is to randomly cast a subset of features multiple times to construct a group of Manifoldrons and integrate all their predictions. Such an approach is called feature bagging whose effectiveness has been confirmed in anomaly detection \cite{zhao2017ensemble}. 
Clearly, feature bagging can exponentially enhance the computational efficiency of the Delaunay triangulation and sample prediction via overcoming the curse of dimensionality and facilitating parallel computation. Table \ref{tab:complexity} summarizes the computational saving due to feature bagging. Other than these advantages, we argue that because of circumventing the high dimensionality, feature bagging is also instrumental to the performance of the Manifoldron in two senses.

On the one hand, Beyer \textit{et al.} \cite{beyer1999nearest} shows that in high dimensional space, samples are less distinguishable from one another for a broad class of data distributions because distances of samples tend to be the same when the dimensionality increases. Consequently, in high dimensional space, not only is the manifold structure not salient but also is the proposed method to profile a manifold doomed. On the other hand, one recent paper \cite{balestriero2021learning} suggests that learning in high dimension always amounts to extrapolation. Supporting this opinion, we also find that as the dimensionality increases, the number of interior test samples of the envelope decreases. 
Since interior samples are more likely to be correctly classified than the exterior ones, using a subset of features somehow enhances the accuracy of the Manifoldron.

\textbf{Point2Point projection.} Since the distance between a test point and a hyperplane is obtained via deriving Eq. \eqref{eqn:distance} slowly, we turn to the distance between a test point and a vertex of an envelope to approximate it in high dimensional space. Accordingly, the closest distance between a point and vertices of an envelope now serves as the distance between a test point and an envelope. Considering that feature bagging has been used, a typical computational complexity of $\mathcal{O}(D'^2)$ for deriving Eq. \eqref{eqn:distance} is reduced to $\mathcal{O}(D')$. The point2point projection is a good approximation to the point2plane projection according to our experiments.

In all, we summarize the time complexity of the Manifoldron in terms of training and inference as follows:
\begin{itemize}
    \item training time: $\mathcal{O}(n+\log n) \sim \mathcal{O}(n^{[D'/2]}+\log n)$ 
    \item inference time: $\mathcal{O}(D'\log n) \sim \mathcal{O}(D'^2\log n) $. 
\end{itemize}

\section{Properties of the Manifoldron}
To put the proposed Manifoldron in perspective, here we systematically investigate the key characteristics of the Manifoldron. Specifically, (i) we analyze the manifold characterization ability of the Manifoldron by proving the universal boundary approximation theorem, and illustrate why the Manifoldron can escape shortcut solutions; (ii) we reveal the connection between the Manifoldron and a neural network by showing that the Manifoldron can be transformed into a neural network with the widespread ReLU activation.

\subsection{Manifold Characterization Capability}
The manifold characterization capabilities of the proposed Manifoldron are twofold: learning complex manifolds and avoiding shortcut solutions.

\textbf{Learning complex manifolds.} We argue that the proposed Manifoldron can learn complicated manifolds better than both parameterized and non-parameterized machine learning models can. On the one hand, a parameterized machine learning model is typically subjected to a certain standard in establishing the decision boundaries, which limits their model expressive ability. For example, a neural network is prescribed to do a hierarchical space partition and tends to have cobweb-like decision boundaries. A decision tree inflexibly divides the space vertically and horizontally. On the other hand, the expressivity of a non-parameterized model is also limited because it does not consider the manifold structure.
In contrast, the proposed Manifoldron directly characterizes a complex manifold by profiling its envelope. Provided sufficient samples, the target manifold can be desirably delineated.  

Mathematically, we prove a universal boundary theorem that given a sufficient number of samples, the Manifoldron can approximate a binary function and its associated decision boundaries arbitrarily well. With this theorem, the expressivity of the Manifoldron should not be worried to some extent. Mathematically, we have the following proposition:

\begin{definition}
Define the function $g: [-B, B]^D \to \{0,1\}$ as a binary function 
\begin{equation}
g(\x) = \Big{\{}\begin{array}{cc}  0,& \quad \ \textbf{if}\quad \x \in \mathcal{R}_0/ \mathcal{H}  \\
1,&\ \textbf{if}\quad \x\in \mathcal{R}_1,
\end{array}
\label{eqn:universal}
\end{equation}
where $\mathcal{R}_0\cup \mathcal{R}_1 = [-B, B]^D$,  interior points of $\mathcal{R}_0$ do not belong to $\mathcal{R}_1$ and vice versa. The decision boundary of $g(\x)$ is $\mathcal{H}(g)=\overline{\mathcal{R}}_0 \cap \overline{\mathcal{R}}_1$.
\end{definition}

\begin{assumption}
The probability density of the data distribution is positive over $[-B, B]^D$. 
\label{no_compact_support}
\end{assumption}

\begin{prop}[universal boundary approximation]
Under the assumption \ref{no_compact_support}, for any function $g(\x)$ defined as Eq. \eqref{eqn:universal}, given the error tolerance $\epsilon_1, \epsilon_2>0$ in estimating $g(\x)$ and the probability tolerance $\delta \in (0,1)$, there exists a number $n^*$ dependent on $\epsilon_1$, $\epsilon_2$, and $\delta$, if the number of training data $n$ satisfies $n \geq n^*$, then with probability at least $1-\delta$, the Manifoldron function $\tilde{g}$ computed based on this training data has the following properties: 
\begin{enumerate}
    \item The decision boundary $\mathcal{H}(\tilde{g})$ of $\tilde{g}$ fulfills $\mathcal{H}(\tilde{g}) \subset \mathcal{H}_{\epsilon_1}(g)$; 
    \item $\tilde{g}$ can approximate $g$ arbitrarily well in terms of $L_2$ norm: 
    $\int_{\x \in [-B,B]^D} ||\tilde{g}-g||^2 d\x \leq \epsilon_2$.
\end{enumerate}
\label{prop:universal_boundary_approximation}
\end{prop}

Our proposition and proof are heavily based on \cite{chen2018explaining}.

\begin{definition}[the relaxed region of a decision boundary]
Given the decision boundary $\mathcal{H}(g)$, we define the $\epsilon$-neighborhood of $\mathcal{H}(g)$ as $\mathcal{H}_\epsilon (g)=\{\x \in [-B, B]^D \ | \ \underset{\x' \in \mathcal{H}(g)}{\min} \|\x-\x'\| \leq \epsilon\}$.
\end{definition}

\begin{definition}
We define a ball centered at $\x$ with a radius $r$ as $\Gamma_{x,r}=\{\x' \in [-B, B]^D \ | \ \|\x-\x'\|\leq r\}$, where $\|\cdot\|$ is the Euclidean norm. 
\end{definition}



\begin{figure}[htb]
\centering
\includegraphics[width=\linewidth]{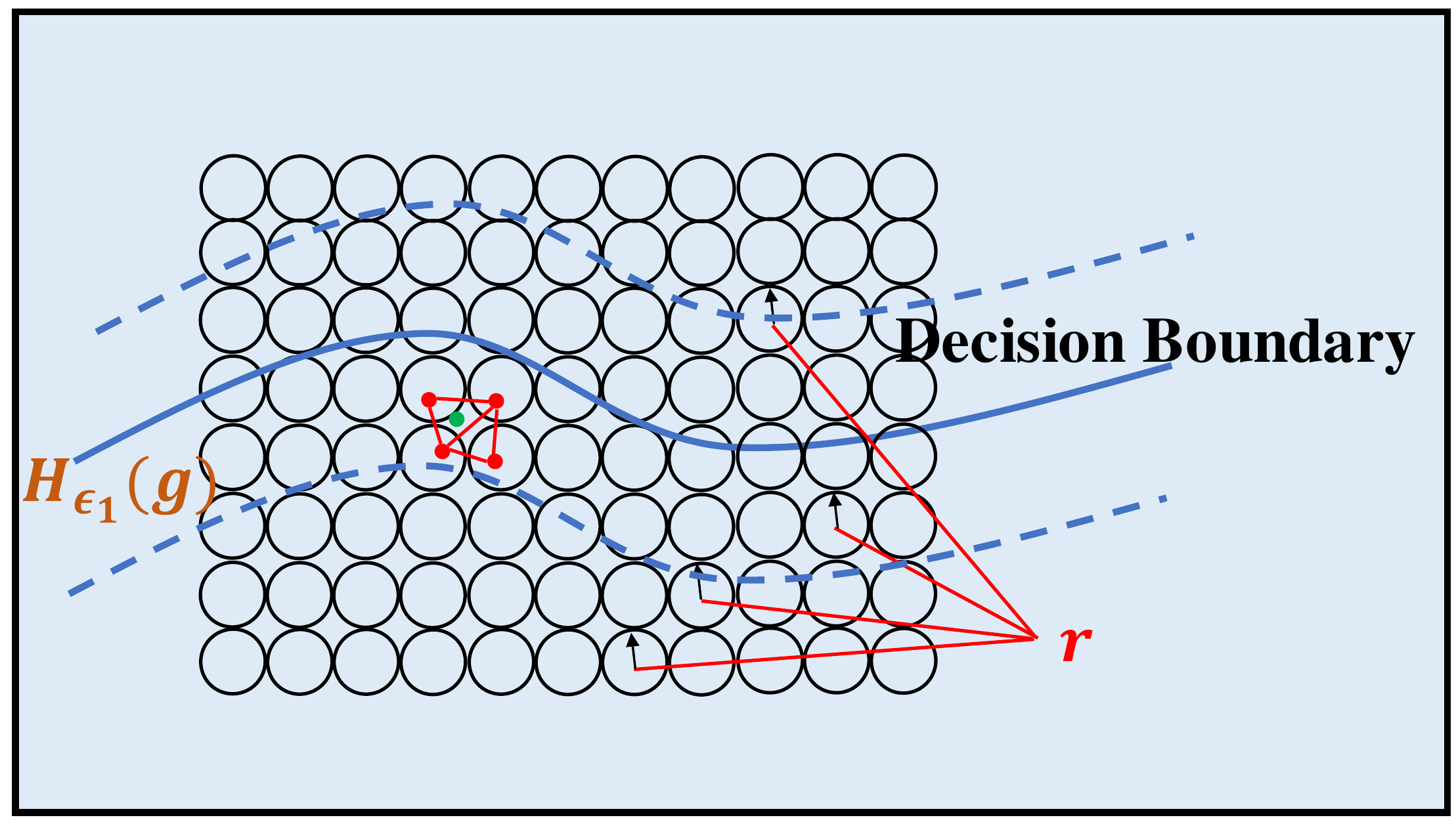}
\caption{The scheme of proving the universal boundary approximation theorem.} 
\label{fig:universal_boundary}
\end{figure}

\begin{lemma}
Given any region $\mathcal{R} \subset [-B,B]^D$ with $\textswab{m}(\mathcal{R})>0$, we i.i.d. sample $n$ points from $[-B,B]^D$ based on the data distribution $\mathbb{P}_{X}$ satisfying the assumption \ref{no_compact_support}, the total number of points landing in $\mathcal{R}$ is $N$, then
\begin{equation}
\mathbb{P}(N>1)  \geq 1 - \mathrm{exp}\Big(-\frac{(n\mathbb{P}_{X}(\mathcal{R})-1)^2}{2n\mathbb{P}_{X}(\mathcal{R})}\Big).
\end{equation}
\label{lem:area_chernoff}
\end{lemma}

\begin{figure*}[htb]
\centering
\includegraphics[width=\textwidth]{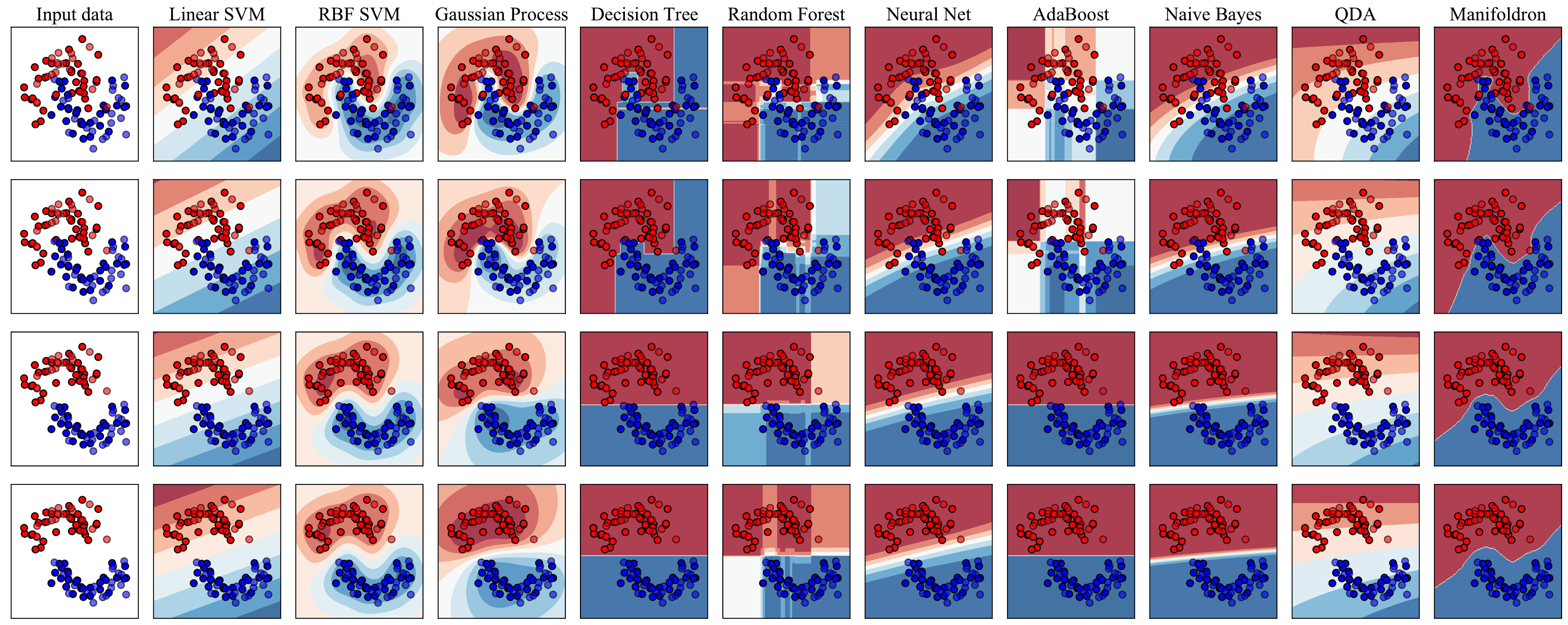}
\caption{The decision boundary contours of different models are obtained from the training data (blue and red clusters).
As the blue cluster moves, the shape of the decision boundary of the Manifoldron is not affected too much compared to other models.} 
\label{fig:classifier}
\vspace{-0.4cm}
\end{figure*}

\begin{proof}
Note that whether or not a new point lands in the region $\mathcal{R}$ conforms to a binomial distribution. Accordingly, $N$, as a sum of points landing in $\mathcal{R}$, follows a Binomial distribution. Applying the Chernoff bound for the binomial distribution \cite{chen2018explaining}, we have
\begin{equation}
    \mathbb{P}(N\leq 1) \leq \mathrm{exp}\Big(-\frac{(n\mathbb{P}_{X}(\mathcal{R})-1)^2}{2n\mathbb{P}_{X}(\mathcal{R})}\Big),
\label{eqn:sample_in_ball}    
\end{equation}
which concludes our proof.
\end{proof}

The above lemma shows that regardless of how tiny a region can be, as long as points landing on it has a non-zero probability, when we sample sufficiently many points, there will be at least one point landing on it.

\begin{proof}
As Figure \ref{fig:universal_boundary} shows, given grid points in $[-B,B]^D$, where two neighbors have a distance of $2r>0$, we can generate regularly arranged balls $\{\Gamma_{i}\}_{i=1}$ of radius $r$ for every point. Later, we show how to determine $r$. Some balls are split by decision boundaries. Let $\mathcal{R}_{0,i}=\Gamma_{i}\cap \mathcal{R}_0$ and $\mathcal{R}_{1,i}=\Gamma_{i}\cap \mathcal{R}_1$. Based on \textbf{Lemma \ref{lem:area_chernoff}}, as long as $\mathcal{R}_{0,i}, \mathcal{R}_{1,i} \neq \varnothing $, with close-to-1 probability, at least one sample will appear in each $\mathcal{R}_{1,i}$ and $\mathcal{R}_{0,i}$. 

For samples of each class, given $2^D$ neighboring balls of one sample (four neighboring balls in 2D space, as shown in Figure \ref{fig:universal_boundary}), there will be at least one sample contained by each ball. Because Delaunay triangulation makes sure that there are no isolated samples, and any region is covered by a triangle, samples whose associated balls are horizontally or vertically adjacent will be connected in Delaunay triangulation. Take the 2D case as an example, four points in Figure \ref{fig:universal_boundary} must be connected horizontally or vertically; otherwise, there will be regions missed by any triangle.  
In the trimming stage, we prescribe that two samples, with a distance of no more than $4r$, are neighbors of each other. In this light, all edges are local. When we derive the envelope, these local edges are connected 
to constitute a part of the envelope within $\mathcal{H}_{4r} (g)$. Based on the envelope, data are labelled with zero or one.

As a result, we get a prediction function $\tilde{g}$ whose decision boundary $\mathcal{H}(\tilde{g}) \subset \mathcal{H}_{4r} (g)$. Let $r \leq \frac{\epsilon_1}{4}$, we prove the property 1. 
Furthermore, the prediction of $\tilde{g}$ is correct in $\mathcal{R}_1/\mathcal{H}_{4r}$ and $\mathcal{R}_0/\mathcal{H}_{4r}$, we have

\begin{equation}
\begin{aligned}
 & \int_{\x \in [-B,B]^D} ||\tilde{g}-g||^2 d\x \\
 = & \int_{\x \in \mathcal{R}_1/\mathcal{H}_{4r}} ||\tilde{g}-g||^2 d\x + \int_{\x \in \mathcal{H}_{4r}} ||\tilde{g}-g||^2 d\x & \\
 & + \int_{\x \in \mathcal{R}_1/\mathcal{H}_{4r}} ||\tilde{g}-g||^2 d\x 
  + \int_{\x \in \mathcal{H}_{4r}} ||\tilde{g}-g||^2 d\x \\
 = & \int_{\x \in \mathcal{H}_{4r}} ||\tilde{g}-g||^2 d\x + \int_{\x \in \mathcal{H}_{4r}} ||\tilde{g}-g||^2 d\x \\
 \leq & \int_{\x \in \mathcal{H}_{4r}} 1 d\x + \int_{\x \in \mathcal{H}_{4r}} 1 d\x \\
 = & C\times(8r).
\end{aligned}
\end{equation}
Here $C$ is a constant. Let $r\leq \frac{\epsilon_2}{8C}$, the property 2 holds true: 
\begin{equation}
    \int_{\x \in [-B,B]^D} ||\tilde{g}-g||^2 d\x \leq \epsilon_2.
\end{equation}
Since we require at least one sample appears in each $\mathcal{R}_{1,i}$ and $\mathcal{R}_{0,i}$, suppose that $\mathbb{P}_X(\mathcal{R}_{0,i})\ \mathrm{and} \  \mathbb{P}_X(\mathcal{R}_{1,i}) \leq p(r)$ which is the function of radius, combining Eq. \eqref{eqn:sample_in_ball}, we have  
\begin{equation}
    \exp\Big(-\frac{(np(r)-1)^2}{2np(r)}\Big) \leq \delta,
\end{equation}
which ends up with 
\begin{equation}
   n > n^*=\frac{2(1+\log(1/\delta))}{p(r)}.
\end{equation}
\end{proof}

\textbf{Avoiding shortcut solutions.} 
Essentially, the shortcut solution is hard to escape for a parameterized machine learning model, as the shortcut solution is a significant minimizer in the landscape of the loss function. Favorably, the construction of the Manifoldron is intrinsically dominated by the manifolds themselves rather than their spatial occupations. Hence, the shape of decision boundaries enjoys robustness.
As Figure \ref{fig:classifier} suggests, because the envelope of the Manifoldron remains the same as the blue cluster moves, the Manifoldron retains less flattened decision boundaries than other models.

\subsection{Link to a Neural Network}
Here, we show that the Manifoldron can be transformed into a neural network, thereby revealing their connection. Such a connection lays a foundation for future research to synergize the Manifoldron and a neural network. For example, we may use the Manifoldron to initialize a neural network or distill the decision boundaries of a network into the Manifoldron.

\begin{prop} Suppose that the representation of a Manifoldron is $g: [-B,~B]^D \to \{0,1\}$, for any $\delta>0$, there exists a network $\mathbf{H}$ of width $\mathcal{O}[D(D+1)(2^D-1)K]$ and depth $D+1$, where $K$ is the minimum
number of simplices to support $g$, satisfying
\begin{equation}
\m\Big(\{\x~|~g(\x)\neq \mathbf{H}(\x)\}\Big) < \delta,
\end{equation}
where $\m(\cdot)$ is the standard measure in $[-B,~B]^D$. 
\label{prop:connection}
\end{prop}
Proof sketch: \textbf{Proposition \ref{prop:connection}} implies that the Manifoldron can be transformed into a neural network. To prove \textbf{Proposition \ref{prop:connection}}, we show that the proposed Manifoldron is a piecewise constant function over a simplicial structure in \textbf{Lemma \ref{lemma:piecewise_constant}}. As we know, a piecewise linear function over a simplicial complex is a combination of linear functions over a simplex. Combining neural networks constructed in \textbf{Lemma \ref{lemma:nn_simplex}} will result in a network that can express the Manifoldron.

\begin{lemma}
A Manifoldron as a classifier is a piecewise constant function over a simplicial complex.
\label{lemma:piecewise_constant}
\end{lemma}

\begin{proof}
Without loss of generality, assume that the proposed Manifoldron works on a binary classification task. Denote the area encompassed by the envelope $\mathcal{E}_c,\ c\in\{0,1\}$ as $\mathcal{A}_c=\{\p ~|~ \p \ \mathrm{is} \ \mathrm{interior} \ \mathrm{to} \ \mathcal{E}_c\}$. Per the definition of the Manifoldron, the prediction to a new point is divided into two cases:

For $\p \in \mathcal{A}_c$, the proposed Manifoldron $g$ maps it to a label $c\in\{0,1\}$. 

For $\p \notin \mathcal{A}_c$, the prediction to $\p$ is based on the minimum distance of $\p$ to all hyperplanes of $\mathcal{E}_c$: $\underset{h\in \mathcal{E}_c }{\min} \ \zeta^{(h)}(\p)$. Let $\tau_c(\p) = \underset{h\in \mathcal{E}_c }{\min} \ \zeta^{(h)}(\p)$, the region outside $\mathcal{A}_0$ and $\mathcal{A}_1$ is splitted into two sub-regions $\mathcal{B}_0=\{\p ~|~ \tau_0(\p)\leq \tau_1(\p)\}$ and $\mathcal{B}_1=\{\p ~|~ \tau_0(\p)\geq \tau_1(\p)\}$. 

Therefore, $g$ is a piecewise constant function:
\begin{equation}
g(\p) = \Big{\{}\begin{array}{cc}  0,& \textbf{if}\quad  \p \in \mathcal{A}_0 \cup \mathcal{B}_0 \\
1,&\ \textbf{if}\quad \p \in \mathcal{A}_1 \cup \mathcal{B}_1.
\end{array}
\label{eqn:manifoldron_function}
\end{equation}
$\mathcal{B}_0 \cap \mathcal{B}_1$ is a piecewise hyperplane where the equidistantly dilated envelopes of $\mathcal{A}_0$ and $\mathcal{A}_1$ intersect. Plus $\mathcal{E}_0$  and $\mathcal{E}_1$ are also piecewise hyperplanes, $\mathcal{B}_0$ and $\mathcal{B}_1$ are simplicial structures, since they are surrounded by $\mathcal{B}_0 \cap \mathcal{B}_1$, $\mathcal{E}_0$  and $\mathcal{E}_1$. Consequently, $\mathcal{A}_0\cup\mathcal{A}_1\cup\mathcal{B}_0\cup\mathcal{B}_1$ is also a simplicial complex. Thus, $g$ is a piecewise constant function over a simplicial complex.
\end{proof}

\begin{lemma}
Suppose that $f: [-B,~B]^D\rightarrow\real$ supported on $\mathrm{S}$ is provided as
\begin{equation}
f(\x) = \Big{\{}\begin{array}{cc}  \mathbf{a}^{\top}\x+b& \textbf{if}\quad  \x \in \mathrm{S}\\
0&\ \textbf{if}\quad \x\in \mathrm{S}^{c},
\end{array}
\end{equation} where $\mathrm{S}^{c}$ is the complement of $\mathrm{S}$, for any $\delta>0$, there exists a network $\mathbf{N}$ of width $\mathcal{O}[D(D+1)(2^D-1)]$ and depth $D+1$, satisfying
\begin{equation}
\m\Big(\{\x~|~f(\x)\neq \mathbf{N}(\x)\}\Big) < \delta,
\end{equation}
where $\m(\cdot)$ is the standard measure in $[-B,~B]^D$. 
\label{lemma:nn_simplex}
\end{lemma}

\begin{proof}
The proof of this lemma is heavily based on \cite{fan2020quasi}. We include results in \cite{fan2020quasi} here for self-sufficiency.

A $D$-simplex $\mathrm{S}$ is a $D$-dimensional convex hull provided by convex combinations of  $D+1$ affinely independent vectors $\{\v_i\}_{i=0}^D \subset \real^D$. We write $V = (\v_1 - \v_0,\v_2 - \v_0,\cdots,\v_D - \v_0)$, then $V$ is invertible, and $S = \left\{\v_0 + V\x ~|~ \x \in \Delta \right\}$ where $\Delta = \left\{\x\in\real^D~|~ \x \geq 0, \mathbf{1}^\top \x \leq 1 \right \}$ is a template simplex in $\real^D$. It is clear that there exists the following one-to-one affine mapping between $\mathrm{S}$ and $\Delta$:  
\begin{equation}
    T:S\rightarrow \Delta, \p \mapsto T(\p) = V^{-1} (\p - \v_0).
\end{equation}
Therefore, we only need to prove the statement on the special case that $S = \Delta$.

We denote the domain of a network as $\Omega = [-B,B]^D$. Given a linear function $\ell(\x) = c_1x_1 + c_2x_2 +\cdots+c_nx_n+ c_{n+1}$, we write $\ell^{-} = \{\x\in\real^D~|~ \ell(\x) <0\}$ 
and $\ell^{+} = \{\x\in\real^D~|~ \ell(\x) \geq 0\}$. $\mathrm{S}$ is enclosed by $D+1$ hyperplanes provided by $\ell_i(\x)=x_i, i = 1,\cdots,D $, and $\ell_{D+1}(\x)=-x_1 -\cdots-x_D +1$. We write $D+1$ vertices of $\mathrm{S}$ as $\v_0 = (0,0,\cdots,0), \v_1 = (1,0,\cdots,0), \v_2 = (0,1,\cdots,0), \cdots, \v_{D+1} = (0,\cdots,0,1)$. Then $f: [-B,~B]^D\rightarrow\real$ supported on $\mathrm{S}$ is provided as
\begin{equation}
f(\x) = \Big{\{}\begin{array}{cc}  \mathbf{a}^{\top}\x+b,& \textbf{if}\quad  \x \in S\\
0,&\ \textbf{if}\quad \x\in S^{c}
\end{array}.
\label{eqn:NDf}
\end{equation}

Our goal is to approximate the given piecewise linear function $f$ using ReLU networks. Let $\sigma(\cdot)$ be a ReLU activation function. We first index the polytopes separated by $D+1$ hyperplanes $\ell_i(\x) = 0, i=1,\cdots,D+1$ as $\mathcal{A}^{(\chi_1, \cdots,\chi_i,\cdots, \chi_{D+1})}=\ell_1^{\chi_1} \cap \cdots \cap \ell_i^{\chi_i} \cap \cdots \cap \ell_{D+1}^{\chi_{D+1}}, \chi_i \in \{+,-\}, i=1,\cdots,D+1$. It is clear to see that $S = \A^{(+,+,\cdots,+)}$. In addition, we use $\vee$ to denote exclusion of certain component. For instance, $\mathcal{A}^{(\chi_1, \vee, \chi_3,\cdots,\chi_{D+1})} =  \ell_1^{\chi_1}  \cap \ell_3^{\chi_3}\cap \cdots \cap \ell_{D+1}^{\chi_{D+1}}$. It can be easily verified that
\begin{equation}
\mathcal{A}^{(\chi_1, \vee, \chi_3,\cdots,\chi_{D+1})} = \mathcal{A}^{(\chi_1, +, \chi_3,\cdots,\chi_{D+1})}\cup \mathcal{A}^{(\chi_1, -, \chi_3,\cdots,\chi_{D+1})}.
\label{eqn:Combine}
\end{equation}
Please note that $\mathcal{A}^{(-,-,\cdots,-)} = \emptyset$. Thus, $D+1$ hyperplanes create a total of $2^{D+1}-1$ polytopes in the $\Omega$. 

Now we recursively define an essential building block, a D-dimensional fan-shaped ReLU network $F_D (\x)$:
\begin{equation}
    \begin{cases}
    F_1(\x) = h_1(\x)\\
    F_{j+1}(\x) = \sigma \circ ( F_j(\x)-\mu^j \sigma \circ h_{j+1}(\x)), \\
    \ \ \ \ \ \ \ \ \ \ \ \ \ \ \ \ \quad j = 1,\cdots, D-1,
    \end{cases}
\end{equation}
where the set of linear functions $\{h_k (\x) =  \p_k^\top \x + r_k\}_{k=1}^{D}$ are provided by $D$ linearly independent vectors $\{\p_k\}_{k=1}^D$, and $\mu$ is a large positive number ($\mu^j$ denotes $\mu$ with the power to $j$). Note that the network $F_D$ is of width $D$ and depth $D$. This network enjoys the following key characteristics: 1) As $\mu \rightarrow \infty$, the hyperplane $h_{1}- \mu h_{2} -\cdots-\mu^j h_{j+1} =0 $ is an approximation to the hyperplane $h_{j+1} =0$ as the term $\mu^j h_{j+1}$ dominates. 
Thus, the support of $F_D (\x)$ converges to $h_1^+ \cap h_2^-\cap \cdots \cap h_D^-$ which is a $D$-dimensional fan-shaped function.
2) Let $C$ be the maximum area of hyperplanes in $[-B,~B]^D$. Because the real boundary $h_{1}- \mu h_{2} -\cdots-\mu^j h_{j+1} =0 $ is almost parallel to the ideal boundary $h_{j+1} =0$, the measure of the imprecise domain caused by $\mu^j$ is at most $C/\mu^j$, where $1/\mu^j$ is the approximate distance between the real and ideal boundaries. In total, the measure of the inaccurate region in building $F_D (\x)$ is at most $C \sum_{j=1}^{D-1}1/\mu^j\leq C/(\mu - 1) $.
3) The function over $D$-dimensional fan-shaped domain is $h_1$, since $(h_j)^+ = 0, j\geq 2$ over this domain. 

Discontinuity of $f$ in Eq. \eqref{eqn:NDf} is one of the major challenges of representing it using a ReLU network. To tackle this issue, we start from a linear function $\tf(\x) = \a^\top \x + b, \forall \x\in\real^D$, which can be represented by two neurons  $\sigma\circ \tf-\sigma\circ (-\tilde{f})$. The key idea is to eliminate $f$ over all $2^{D+1}-2$ polytopes outside $\mathrm{S}$ using the $D$-dimensional fan-shaped functions.

Let us use $\mathcal{A}^{(+,+,+,-,\cdots,-)}$ and $\mathcal{A}^{(+,+,-,-,\cdots,-)}$ to show how to cancel the function $\tf$ over the polytopes outside $\mathrm{S}$. According to Eq. \eqref{eqn:Combine}, $\mathcal{A}^{(+,+,+,-,\cdots,-)}$ and $\mathcal{A}^{(+,+,-,-,\cdots,-)}$ satisfy
\begin{equation}
    \mathcal{A}^{(+,+,\vee,-,\cdots,-)} = \mathcal{A}^{(+,+,+,-,\cdots,-)} \cup \mathcal{A}^{(+,+,-,-,\cdots,-)},
\label{eq:neighbors}    
\end{equation}
where $\mathcal{A}^{(+,+,\vee,-,\cdots,-)}$ is a $D$-dimensional fan-shaped domain. Without loss of generality, a number $D+1$ of $D$-dimensional fan-shaped functions over $\mathcal{A}^{(+,+,\vee,-,\cdots,-)}$ are needed as the group of linear independent bases to cancel $\tf$ , where the $k^{th}$ fan-shaped function is constructed as
\begin{equation}
    \begin{cases}
        & F_1^{(k)} = x_1-\eta_k x_k\\
        & F_2^{(k)} = \sigma \circ \big(F_1^{(k)}-\mu \sigma \circ (-x_2) \big)\\        
        & F_{3}^{(k)} =  \sigma \circ\big(F_{2}^{(k)} -\mu^2 \sigma \circ(x_{4}) \big) \\
        & F_{4}^{(k)} = \sigma \circ \big(F_{3}^{(k)} -\mu^3 \sigma \circ(x_{5}) \big) \\
        & \ \ \ \ \vdots \\
        & F_{D}^{(k)} =   \sigma \circ \big(F_{D-1}^{(k)}-\mu^{D-1} \sigma \circ(-x_1-\cdots-x_D+1) \big),
    \end{cases}
\end{equation}
where we let $x_{D+1}=1$ for consistency, the negative sign for $x_2$ is to make sure that the fan-shaped region $\ell_1^+ \cap (-\ell_2)^- \cap \ell_4^- \cap \cdots \cap \ell_{D+1}^-$ of $F_{D}^{(k)}$ is $\mathcal{A}^{(+,+,\vee,-,\cdots,-)}$, $\eta_1=0$, and $\eta_k=\eta, k=2,...,D+1$ represents a small shift for $x_1=0$ such that $\m((x_1)^+ \cap (x_1-\eta_k x_k)^-) < C\eta_k $. The constructed function over $\mathcal{A}^{(+,+,\vee,-,\cdots,-)}$ is
\begin{equation}
F_{D}^{(k)} = x_1-\eta_k x_k, k=1,\cdots,D+1,
\end{equation}
which is approximately over 
\begin{equation}
    \forall \x\in\mathcal{A}^{(+,+,\vee,-,\cdots,-)}\backslash \Big((x_1)^+ \cap (x_1-\eta_k x_k)^-) \Big)  .
\end{equation}
Let us find $\omega_1^*,\cdots,\omega_{D+1}^*$ by solving 
\begin{equation}
\begin{bmatrix}
1 &1 & 1 &\cdots &  1\\
0 &-\eta & 0 &\cdots &  0 \\
0 &0 & -\eta &\cdots &  0 \\
\vdots&  &  & \vdots & \\
0 &0 & 0 &\cdots &  -\eta \\
\end{bmatrix} \begin{bmatrix}
\omega_1 \\
\omega_2  \\
\omega_3 \\
\vdots \\
\omega_{D+1}
\end{bmatrix}=-\begin{bmatrix}
a_1 \\
a_2  \\
a_3 \\
\vdots \\
b
\end{bmatrix},
\end{equation}
and then the new function 
$F^{(+,+,\vee,-,\cdots,-)}(\x)=\sum_{k=1}^{D+1}\omega_k^* F_D^{(k)}(\x)$ satisfies that
\begin{equation}
\begin{aligned}
       &  \m\Big(\{\x\in \mathcal{A}^{(+,+,\vee,-,\cdots,-)}|~\tf(\x) + F^{(+,+,\vee,-,\cdots,-)}(\x) \neq 0 \} \Big) \\
     & \leq C (D\eta+\frac{D+1}{\mu-1}).
\end{aligned}
\end{equation}

Similarly, we can construct other functions $\overbrace{F^{(+,-,\vee,-,\cdots,-)}(\x),F^{(+,\vee,-,-,\cdots,-)}(\x),\cdots}^{~2^{D}-2 \ \ terms} $ 
to cancel $\tf$ over other polytopes. Finally, these $D$-dimenional fan-shaped functions are aggregated to form the following wide ReLU network $\mathbf{N}(\x)$:
\begin{equation}
\begin{aligned}
    & \mathbf{N} (\x) = \sigma\circ (\tf(\x))-\sigma\circ (-\tf(\x)) \\
    & +\overbrace{F^{(+,+,\vee,-,\cdots,-)}(\x) + F^{(+,\vee,-,-,\cdots,-)}(\x)+\cdots}^{~{2^{D}-1} \ \ terms},
\end{aligned}    
\end{equation}
where the width and depth of the network are  $D(D+1)(2^{D}-1)+2$ and $D+1$ respectively. In addition, because there are $2^{D}-1$ polytopes being cancelled, the total area of the regions suffering from errors is no more than  
\begin{equation}
    (2^{D}-1)C (D\eta+\frac{D+1}{\mu-1}).
\end{equation}
Therefore, for any $\delta>0$, as long as we choose appropriate $\mu, \eta $ that fulfill
\begin{equation}
\begin{aligned}
0 < 1/{(\mu-1)} , \eta &< \frac{\delta}{(2^{D}-1)C(D+D+1)} \\
&=\frac{\delta}{(2^{D}-1)C(2D+1)},
\end{aligned}
\end{equation}
the constructed network $\mathbf{N}(\x)$ will have
\begin{equation}
    \m\left(\{\x\in\real^D~|~f(\x)\neq \mathbf{N}(\x)\}\right) < \delta.
\end{equation}

\end{proof}

\noindent\textit{Proof of Proposition \ref{prop:connection}.}
A piecewise constant function is a special case of a piecewise linear function, therefore, we only need to work on the piecewise linear function. Since a piecewise linear function over a simplicial structure can be decomposed into linear functions over a simplex, we can aggregate the network $\mathbf{N}(\x)$ concurrently to obtain a meta-network:
 \begin{equation}
 \mathbf{H}(\x) = \sum_{k=1}^K \mathbf{N}^{(k)}(\x),
 \end{equation}
where $\mathbf{N}^{(k)}(\x)$ represents the linear function over the $k^{th}$ simplex, and $K$ is the minimum number of needed simplices to support $g$ . Therefore, the constructed wide network $\mathbf{H}(\x)$ is of width $\mathcal{O}[D(D+1)(2^D-1)K]$ and depth $D+1$. $\hfill\square$

\section{Synthetic Experiments}
\begin{figure}[htb]
\centering
\includegraphics[width=\linewidth]{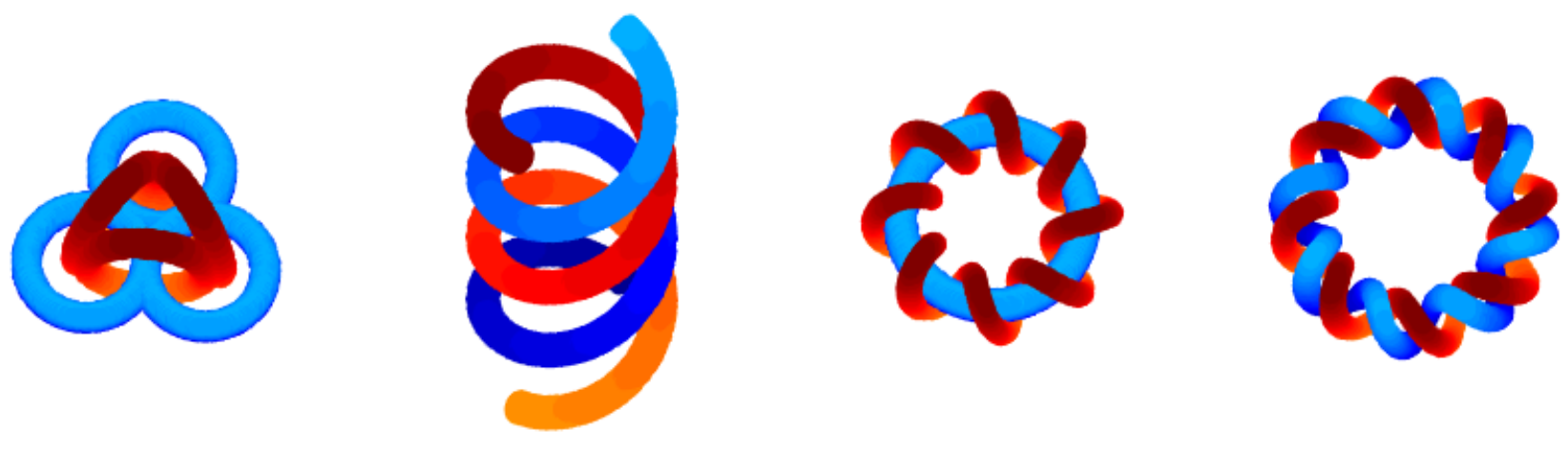}
\caption{The constructed 3D manifolds for the Manifoldron and other models to classify. } 
\label{fig:fancy3D}
\end{figure}

As shown in Figure \ref{fig:fancy3D}, we construct four complicated manifolds for the Manifoldron and other models to classify, thereby verifying the manifold characterization ability of the Manifoldron. From left to right, for convenience we call these manifolds \textit{SixCircles}, \textit{DNA}, \textit{RingSpiral}, and \textit{TwoCurlySpirals}, respectively. Each manifold consists of two submanifolds representing two classes, denoted as blue and red classes for simplicity.

\textbf{3D manifold generation.}
Four complicated manifolds are constructed via mathematical functions. The main idea to construct them is to move a circular face along a 3D trajectory while keeping the normal vector of the surface parallel to the tangent vector of the trajectory. The 3D region the circular face sweeps constitutes the target manifold. 

\begin{itemize}
    
\item \textit{SixCircles}: The blue class consists of three circular trajectories whose parametric equations are 
\begin{equation}
\begin{cases}
    & x_1 = 0.4\cdot \cos(t)\\
    & y_1 = 0.4\cdot \sin(t) \\
    & z_1 = 0 \\
\end{cases}
\quad\text{,}\quad
\begin{cases}
    & x_2 = 0.8+0.4\cdot \cos(t)\\
    & y_2 = 0.4\cdot \sin(t) \\
    & z_2 = 0 \\
\end{cases},
\end{equation}
and
\begin{equation}
\begin{cases}
    & x_3 = 0.45+0.4\cdot \cos(t)\\
    & y_3 = 0.7+0.4\cdot \sin(t) \\
    & z_3 = 0 \\
\end{cases}
\end{equation}
to guide the movement of the circular face. The red class consists of three circular trajectories, and rotating the one trajectory can obtain the rest two. Therefore, we just show the parametric equation of the simplest trajectory that is parallel to the y-axis: 
\begin{equation}
\begin{cases}
    & x_4 = 0.4\cdot \cos(t)\\
    & y_4 = 0.4 \\
    & z_4 = 0.4\cdot \sin(t) \\
\end{cases}.
\end{equation}

\item \textit{DNA}: The trajectory to generate data in blue class is a spiral whose parametric equation is 
\begin{equation}
\begin{cases}
    & x_1 = 0.2\cdot \cos(t)\\
    & y_1 = 0.2\cdot \sin(t) \\
    & z_1 = 0.1\cdot t \\
\end{cases},
\end{equation}
and the trajectory used in the red class is $\pi$ later than the one in the blue class:
\begin{equation}
\begin{cases}
    & x_2 = 0.2\cdot \cos(t-\pi)\\
    & y_2 = 0.2\cdot \sin(t-\pi) \\
    & z_2 = 0.1\cdot t \\
\end{cases}.
\end{equation}

\begin{table}[ht]
\caption{Statistics of datasets made by complicated manifolds. }
\centering
\scalebox{0.85}{
\begin{tabular}{|l|c|c|c|}
\hline
    Datasets & \#Train Instances & \#Test Instances & \#Classes  \\ 
    \hline
    SixCircles  &   6732 & 726 & 2 \\   
    \hline
    DNA & 8484 & 3232 & 2 \\
    \hline
    RingSpiral & 11044 & 5522 & 2 \\
    \hline
    TwoCurlySpirals & 9624 & 4872 & 2 \\
     \hline
\end{tabular}}
\label{tab:manifold_statistics}
\end{table}

\item \textit{RingSpiral}: In this construction, the blue class is a ring whose parametric function is similar to the aforementioned ones:
\begin{equation}
\begin{cases}
    & x_1 = \cos(t)\\
    & y_1 = \sin(t) \\
    & z_1 = 0 \\
\end{cases}.
\end{equation}
The red class is a curly spiral entangling with the ring, which can be described as
\begin{equation}
\begin{cases}
    & x_2 = \cos(t)\cdot(0.3\cdot \sin(8\cdot t)+1)\\
    & y_2 = \sin(t)\cdot(0.3\cdot  \sin(8\cdot t)+1) \\
    & z_2 = 0.3\cdot \cos(8\cdot t) \\
\end{cases}.
\end{equation}

\item \textit{TwoCurlySpirals}: Replacing the ring in the above with another spiral, parametric equations of two trajectories here are
\begin{equation}
\begin{cases}
    & x_1 = \cos(t)\cdot(0.3\cdot \sin(8\cdot t)+1)\\
    & y_1 = \sin(t)\cdot(0.3\cdot  \sin(8\cdot t)+1) \\
    & z_1 = 0.3\cdot \cos(8\cdot t) \\
\end{cases}
\end{equation}
and
\begin{equation}
\begin{cases}
    & x_1 = \cos(t-0.1\cdot\pi)\cdot(0.3\cdot \sin(8\cdot(t-0.1\cdot\pi))+1)\\
    & y_1 = \sin(t-0.1\cdot\pi)\cdot(0.3\cdot  \sin(8\cdot(t-0.1\cdot\pi))+1) \\
    & z_1 = 0.3\cdot \cos(8\cdot(t-0.1\cdot\pi)) \\
\end{cases}.
\end{equation}

\end{itemize}

\begin{table*}[ht]
\caption{Quantitative classification results of different models on small and large datasets. The best performance on each dataset is bold-faced, and * remarks tied winners.}
\centering
\scalebox{0.85}{
\begin{tabular}{|l|c|c|c|c|c|c|c|c|c}
    \hline
    Datasets &  Linear-SVM & RBF-SVM & DT  & RF & NN&  AdaBoost  &  QDA  & Manifoldron \\ 
    \hline
    SixCircles  & 0.4697 & \textbf{1.0}* & 0.8939 & 0.9091 & 0.8636 & 0.8939  & 0.8636 & \textbf{1.0}* \\    
    \hline
    DNA  & 0.4805 & 0.9895 & 0.3453 & 0.7704 & 0.6470 & 0.4678  & 0.5473 & \textbf{1.0} \\
    \hline
    RingSpiral  & 0.7988 &  \textbf{1.0}* & 0.8544 & 0.8745 & 0.7988 & 0.9596  & 0.7983 & \textbf{1.0}*\\
    \hline
    TwoCurlySpirals  & 0.4759& 0.7238 & 0.5914 & 0.7598 & 0.5 & 0.5736  & 0.4991 & \textbf{1.0}\\
    \hline
\end{tabular}}

\label{tab:quant_manifolds}
\end{table*}

\textbf{Dataset curation.} Let $r$ be the radius of a circle in the circular face, given $r_1 < r_2 < r_3$, we use the circle with the radius of $r_1$ and $r_3$ to generate the training data and the circle with the radius of $r_2$ to generate the test data. By doing so, the test data are totally surrounded by the training data, and there is no outliers in the test data.

\begin{itemize}
   \item \textit{SixCircles}: $r_1=0.01, r_2=0.02, r_3=0.04$. We sample $t$ from $[0,2\pi]$ with a step of $0.04\pi$ for training data and $0.2\pi$ for test data.

\item \textit{DNA}: $r_1=0.01, r_2=0.02, r_3=0.04$. We sample $t$ from $[0,4\pi]$ and $[\pi, 5\pi]$ for two spirals with a step of $0.04\pi$ for the training and test data, respectively.

\item \textit{RingSpiral}: $r_1=0.01, r_2=0.02, r_3=0.05$. We sample $t$ from $[0,2\pi]$ for the ring with a step of $0.02\pi$, and we cast $t$ from $[0,8\pi]$ for the spiral with a step of $0.02\pi$.
   
\item \textit{TwoCurlySpirals}: $r_1=0.01, r_2=0.02, r_3=0.05$.
We sample $t$ from $[0,8\pi]$ and $[\pi, 9\pi]$ with a step of $0.02\pi$ for the training and test data, respectively.

\end{itemize}
Per the above protocols, we curate four datasests whose stastitics are summarized in Table \ref{tab:manifold_statistics}.

\textbf{Classification results.} We compare the Manifoldron with parametrized machine learning algorithms: SVM with a linear kernel (Linear-SVM), SVM with a radial basis function (RBF) kernel (RBF-SVM), DT, RF, NN, AdaBoost, and QDA. We use the classification accuracy as a performance metric. Table \ref{tab:quant_manifolds} shows classification results of all algorithms. It can be seen that the proposed Manifoldron achieves the highest accuracy for all four datasets, whereas the RBF-SVM model is the tied winner on the \textit{SixCircles} and \textit{RingSpiral} datasets. The reason why the RBF-SVM model performs well on this two datasets is that the radial basis function transform fits the ring manifolds there, \textit{i}.\textit{e}., transforming a ring into a concentrated cluster to facilitate classification. Moreover, the advantages of the Manifoldron on \textit{DNA} and \textit{TwoCurlySpirals} are considerable relative to its competitors. Indeed, the classifiers such as Linear-SVM, DT, NN, and QDA are not good at distinguishing curly and interwined manifolds. In contrast, the Manifoldron excels at this job because it deals with the intrinsic manifold. Regardless of how complicated a manifold is, provided sufficient data, the envelopes can be correctly identified to make the decision boundaries.

\section{Public Benchmark Experiments}
In this experiment, we evaluate the proposed model and address the following questions: i) Can the Manifoldron deliver the competitive classification performance? ii) Can the Manifoldron scale to the large-scale datasets? We remark that the Manifoldron is a machine learning model with an
emphasis on the tabular data instead of computer vision. Therefore, we conduct experiments on tabular datasets. 

\textbf{Datasets.} We conduct experiments on 9 small and 11 large well-known datasets in machine learning, which are the same with recent studies published in prestigious venues \cite{wang2021scalable, wang2020transparent, yang2017scalable}.  Table \ref{tab:statistics} summarizes the statistics of these datasets.

\textbf{Code Disclosure and Data Source}
All datasets are publicly available from python scikit-learn package, UCI machine learning repository, Kaggle, and Github. The links to access all utilized datasets are \textit{circle}\footnote{https://scikit-learn.org/stable/modules/generated/sklearn.datasets.make\_circles.html}, \textit{glass}\footnote{https://archive.ics.uci.edu/ml/datasets/glass+identification}, \textit{ionosphere}\footnote{https://archive.ics.uci.edu/ml/datasets/ionosphere}, \textit{iris}\footnote{https://scikit-learn.org/stable/auto\_examples/datasets/plot\_iris\_dataset.html},  \textit{moons}\footnote{https://scikit-learn.org/stable/modules/generated/sklearn.datasets.make\_moons.html}, \textit{parkinsons}\footnote{https://archive.ics.uci.edu/ml/datasets/parkinsons}, \textit{seeds}\footnote{https://archive.ics.uci.edu/ml/datasets/seeds}, \textit{spirals}\footnote{https://github.com/Swarzinium-369/Two\_Spiral\_Problem}, \textit{wine}\footnote{https://scikit-learn.org/stable/modules/generated/sklearn.datasets.load\_wine.html}, 
\textit{banknote}\footnote{https://archive.ics.uci.edu/ml/datasets/banknote+authentication},
\textit{breast}\footnote{https://archive.ics.uci.edu/ml/datasets/breast+cancer+wisconsin+(diagnostic)},
\textit{chess}\footnote{https://archive.ics.uci.edu/ml/datasets/Chess+(King-Rook+vs.+King-Pawn)},
\textit{drug}\footnote{https://archive.ics.uci.edu/ml/datasets/Drug+consumption+\%28quantified\%29},
\textit{letRecog}\footnote{https://archive.ics.uci.edu/ml/datasets/Letter+Recognition},
\textit{magic04}\footnote{https://archive.ics.uci.edu/ml/datasets/magic+gamma+telescope},
\textit{nursery}\footnote{https://archive.ics.uci.edu/ml/datasets/nursery},
\textit{satimage}\footnote{https://archive.ics.uci.edu/ml/datasets/Statlog+(Landsat+Satellite)},
\textit{semeion}\footnote{https://archive.ics.uci.edu/ml/datasets/semeion+handwritten+digit},
\textit{tic-tac-toe}\footnote{https://archive.ics.uci.edu/ml/datasets/Tic-Tac-Toe+Endgame},
\textit{usps5}\footnote{https://www.kaggle.com/bistaumanga/usps-dataset}. The statistics of these datasets are summarized in Table \ref{tab:statistics}.

\textbf{Experimental setups.} We compare the Manifoldron with 11 mainstream machine learning algorithms: KNN \cite{keller1985fuzzy}, SVM with a linear kernel (Linear-SVM), SVM with a radial basis function (RBF) kernel (RBF-SVM) \cite{hearst1998support}, C4.5 \cite{quinlan2014c4}, RF \cite{breiman2001random}, NN \cite{lecun2015deep}, AdaBoost \cite{schapire2013explaining}, Naive Bayes \cite{rish2001empirical}, QDA \cite{srivastava2007bayesian}, XGBoost \cite{chen2016xgboost} and LightGBM \cite{ke2017lightgbm}. We use the classification accuracy as a performance metric. Moreover, we compute the average rank to holistically evaluate different models. Since data from different classes pertain to different manifolds, the training data in the Manifoldron are dealt class by class. 
Sometimes, (Delaunay) triangulation fails because features of partial data are linearly dependent. To fix this problem, minor disturbance is added to undermine such a dependence.
In the trimming stage, we set the neighbor number of each sample to 14 by default in querying the neighbors from the KD tree. In the test stage, we adopt feature bagging to reduce the computational complexity of the Manifoldron and improve the performance. We randomly sample four to seven features from the entire feature space each time without replacement. All predictions adopt the most frequent class as the final prediction. We randomly split all datasets into 70\% for training and 30\% for test. Each model runs five times to get a solid result. 
The experiments are implemented in Python 3.7 on Windows 10 using CPU Intel i7-8700H processor at 3.2GHz. 

\begin{table}[h]
\caption{Statistics of all datasets. }
\centering
\scalebox{0.85}{
\begin{tabular}{l|c|c|c|c}
\hline
    \hline
    Datasets & \#Instances & \#Classes & \#Features & Feature Type \\ 
    \hline
    circle  &   100 & 2 & 2 & continuous\\   
    \hline
    glass& 214 & 7 & 10 & continuous\\
    \hline
    ionosphere & 351 & 2 & 34 &  mixed \\
     \hline
     iris & 150 & 3 & 4 & continuous  \\
    \hline
    moons  &   100 & 2 & 2 & continuous\\
    \hline
    parkinsons & 197 & 2 & 23 & continuous \\
    \hline
    seeds & 210 & 3 & 7 & continuous  \\
     \hline
    spirals &   400 & 2 & 2 & continuous  \\
    \hline
    wine & 178 & 3 &  13 & continuous \\
    \hline
     \hline
     banknote & 1372 & 2 & 4 & continuous \\
    \hline
     breast & 569 & 2 & 30 & continuous \\
     \hline
     chess & 3195 & 2 & 36 & discrete \\
     \hline
     drug & 1885 & 6 & 32 & continuous \\
     \hline
     letRecog & 20000 & 26 & 16 & discrete\\
     \hline
     magic04 & 19020 & 2 & 10 & continuous \\
     \hline
     nursery &  12960 & 5 & 8 & discrete \\
     \hline
     satimage & 4435 & 6 & 36 & discrete\\
    \hline
     semeion & 1593 & 10 & 256 & discrete\\
     \hline
     tic-tac-toe & 958 & 2 & 9 & discrete \\
     \hline
     usps5 & 5427 & 5 & 256 & discrete\\
     \hline
    \hline
\end{tabular}}
\label{tab:statistics}
\end{table}

\begin{table*}[ht]
\caption{Quantitative classification results of different models on small and large datasets. The best performance on each dataset is bold-faced, and * remarks tied winners. All results are the mean value of five runs.}
\centering
\scalebox{0.85}{
\begin{tabular}{l|c|c|c|c|c|c|c|c|c|c|c|c}
\hline
    \hline
    Datasets & kNN & Linear-SVM & RBF-SVM & C4.5  & RF & DNN&  AdaBoost & Naive Bayes  &  QDA  & XGBoost & LightGBM & Manifoldron \\ 
    \hline
    \hline
    circle  &   0.9625 & 0.4250   & \bf{0.9875}*   & 0.9750 & 0.9750     & 0.9500  & 0.9750  & \bf{0.9875}*  & \bf{0.9875}* & 0.9666 & 0.9773 & \bf{0.9875}* \\    
    \hline
    glass & \bf{1.0} & 0.9688 & 0.9686  & 0.9375& 0.7656 & 0.9843 & 0.7968 & 0.9062 & 0.7968 & 0.9531 & 0.9375 & 0.9843\\
    \hline
    ionosphere & 0.8285 & 0.8476 & 0.8571  & 0.8380 & 0.8857 & 0.9047 & 0.8761 & 0.8285 & 0.8476 & 0.9047 & 0.8952&  \textbf{0.9333}\\
    \hline
    iris & 0.9667 & 0.9333 &\bf{1.0}* & \bf{1.0}*  & 0.933 &\bf{1.0}* & 0.9667 & 0.9667 & \bf{1.0}* & \bf{1.0}* & \bf{1.0}* & \bf{1.0}* \\
    \hline
    moons  &   0.9375 &  0.8125 &  0.9500  &  0.8375 & 0.9250  & 0.8625  & 0.9125 & 0.8625 &  0.8625 & 0.9100 & 0.8967 & \bf{0.9625} \\
    \hline
    parkinsons & 0.8135 & 0.7796 & 0.7457  & 0.7796 & 0.8644 & 0.8305 & 0.8644 & 0.8305 & 0.7796 & \bf{0.8983} & 0.8474 & 0.8813\\
     \hline
    seeds & 0.9524 & 0.9524 & 0.9524  & 0.9524 & 0.9286 & \bf{0.9762}* & 0.7381 & \bf{0.9762}* & \bf{0.9762}* & 0.9047 & 0.9285 & \bf{0.9762}* \\
    \hline
    spiral & \bf{0.9906}* & 0.7625  & 0.9719  & 0.9219  & 0.9594 & 0.9813 & 0.9406 & 0.7563 & 0.7625 & 0.9833 & 0.9875 & \bf{0.9906}*\\
    \hline
     wine & 0.7778 & 0.9444 & 0.8333 & 0.9722 & 0.9444 & 0.9444 & 0.8611 & 0.9167 & \bf{1.0} & 0.9444 & 0.9815 & 0.9629\\
    \hline
    \hline
     banknote & \bf{1.0}* & 0.9709 & \bf{1.0}*  & 0.9782 & 0.9600 & \bf{1.0}* & 0.9964 & 0.8109 & 0.9709 & 0.9964 & 0.9964 & \bf{1.0}*\\
     \hline
     breast & 0.9415 & 0.9707 & 0.9649 &  0.9239 & 0.9532 & \bf{0.9766}* & 0.9239 & 0.9064 &  0.9591 & \bf{0.9766}* & 0.9707 & 0.9415\\
     \hline
     chess & 0.8075 & 0.7856 & 0.5446  & 0.8591 & 0.7918 & 0.8716 & 0.8638 & 0.8043 & 0.8450 & 0.9014 & \bf{0.9030} & 0.7793\\ 
     \hline
     drug & 0.7429 & 0.7712 & 0.7712  & 0.7269 & 0.7712 & 0.7712 & 0.3723 & 0.6737 & 0.7624 & 0.7712 & \bf{0.7730} & 0.7677\\
     \hline
     letRecog & 0.9517 & 0.8335 & 0.8122 & 0.3857 & 0.5817 &  0.8915 & 0.2215 & 0.6262 & 0.8732 & 0.9467 & \bf{0.9532} & 0.9300 \\
     \hline
     magic04 & 0.8354 & 0.7964 & 0.6847 & 0.7983 & 0.8047 & 0.8060 & 0.7947 & 0.7460 & 0.7727 & \bf{0.8875} & 0.8808 & 0.8183\\
     \hline
     nursery &  0.8641 & 0.7841 & 0.9208  & 0.8158 & 0.8275 & 0.9141 & 0.5875 & 0.8383& 0.8541 & 0.9881 & \bf{0.9889} & 0.8458\\
     \hline 
     satimage & 0.6965 & 0.6550 & 0.2305  & 0.6475 & 0.6300 & 0.6280 & 0.5265 & 0.6205 & 0.6420 & 0.9050 & \bf{0.9100} & 0.7030\\
    \hline
    semeion & 0.8526 & 0.8934 & \bf{0.9090}  &  0.4796 & 0.4890 & 0.8652 & 0.2727 & 0.8495 & 0.1222 & 0.8308 & 0.884 & 0.8025\\
     \hline
    tic-tac-toe & 0.7396 & 0.6615 & 0.6615  & 0.8229& 0.7343 & 0.7813 & 0.6927 & 0.6823 & 0.7188 & 0.8923 & \bf{0.8958} & 0.8298 \\
     \hline
     usps5 & 0.9197 & 0.9247 & 0.9048  & 0.6552 & 0.5650 & 0.9262 & 0.7184 & 0.7897 & \bf{0.9362} & 0.9157 & 0.9257 & 0.8644\\
     \hline
    \hline
    avg rank &5.975& 7.825& 6.7  & 8.025& 7.9  & 4.725& 8.875& 8.85 & 6.9  &  4.325 & \textbf{3.55} & 4.35 \\
    \hline
    \hline
\end{tabular}}
\vspace{-0.3cm}
\label{tab:quant}
\end{table*}

\textbf{Hyperparameters of other models.} We set the hyperparameters of other models as follows. 

\begin{itemize}
\item The \textit{k-}nearest neighbor method uses a ball tree with a leaf size 30, a KD tree with a leaf size 30, and the brute-force search strategies to derive 3 nearest neighbors, respectively. It chooses the best predictive result among them. The distance is the Euclidean distance. 

\item For the Linear-SVM, we use a linear kernel with a regularization parameter of 0.025. For the RBF-SVM with RBF kernel, we search the regularization parameter of \{1, 10\} and kernel coefficient of \{0.001, 0.01, 0.1, 1, 10\}. Both SVMs terminate at the error of $10^{-3}$. 

\item The C4.5 is a decision tree model using the Gini impurity \cite{Rai2004gini} to measure the quality of all possible splits and to choose the best one. Each split is mandatory to have at least 2 samples for any internal node. 
\item The random forest classifier encompasses 10 trees whose depth is 5 and feature number is 1 at each split. 

\item We use a one-hidden-layer fully connected neural network of 100 neurons with ReLU activation. The batch size is 200, and the $L_2$ penalty parameter is $10^{-3}$. The Adam algorithm is used to optimize the neural network with a fixed learning rate of 0.001 for 1000 iterations.  

\item The AdaBoost model integrates 50 base estimators (decision trees) into a boosted ensemble. The boosting type is stagewise additive modeling using a
multi-class exponential loss function. 

\item In the naive Bayes method, a perturbation is added to features to increase model stability. The magnitude of the perturbation is no more than $10^{-9}$ of the maximum variance of original features.

\item The Quadratic discriminant analysis estimates the rank of centered matrices of samples in each and every class. In doing so, the significance threshold for a singular value is $10^{-4}$. 

\item The XGBoost uses the gbtree booster with a weight shrinkage of 0.3 and an $\mathrm{L}_2$ regularization term of 1. The maximum depth of a tree is set to 6. A total of 12 threads are used to run the algorithm in parallel. 

\item The LightGBM uses the traditional gradient boosting decision tree as a booster. The number of the leaves in each tree is 31. The number of boosting iterations is 100 with a learning rate of 0.1. 

\end{itemize}

\textbf{Classification performance.} Table \ref{tab:quant} summarizes classification results of all algorithms, where upper rows are for small datasets and lower ones for large datasets. Regarding small datasets, the Manifoldron achieves the highest accuracy for most datasets except for the \textit{glass} and \textit{wine}. Nevertheless, the Manifoldron is only surpassed by the kNN model on the \textit{glass} dataset and takes the third place on the \textit{wine} dataset, only inferior to QDA and DT. As far as large datasets are concerned, the Mainfoldron is the best performer on the \textit{banknote}. For the rest, the Manifoldron still consistently delivers satisfactory performance and enjoys an upper rank compared to its counterparts. For example, on the \textit{letRocog} and \textit{drug} datasets, the gap between scores of the best model and our model is rather marginal. Holistically, the Manifoldron takes a competitive average rank score of \textbf{4.35} across all 20 benchmarks. We conclude that the Manifoldron is superior to or comparable with other mainstream methods. 


\begin{table}[h]
\caption{The training and test time of the Manifoldron on representative large datasets.}
\centering
\scalebox{0.82}{
\begin{tabular}{|l|c|c|c|c|c|c|c|c|}

    \hline

    Datasets &
     \multicolumn{2}{c|}{breast} &
    \multicolumn{2}{c|}{drug} &
    \multicolumn{2}{c|}{letRecog} \\
    \hline
    Type & train & test & train & test & train & test  \\ 
    \hline
    Time  & 1m15s & 0.62s  & 1m30s & 0.23s &  30m4s & 1.72s \\ 
    \hline
    Datasets &
    \multicolumn{2}{c|}{nursery} &
    \multicolumn{2}{c|}{semeion} &
    \multicolumn{2}{c|}{tic-tac-toe}  \\
    \hline
    Type & train & test & train & test & train & test  \\ 
    \hline
    Time  & 5m2s & 0.10s & 37m51s & 13.95s & 31s & 0.02s   \\ 
    \hline
    Datasets &
    \multicolumn{2}{c|}{magic04} &
    \multicolumn{2}{c|}{satimage} &
    \multicolumn{2}{c|}{usps5} \\
    \hline
    Type & train & test & train & test & train & test  \\ 
    \hline
    Time  & 30m13s & 0.40s & 25m7s & 1.26s  & 3h6m54s & 5.64s  \\ 
    \hline    
    
\end{tabular}}
\label{tab:time}
\end{table}

\begin{figure*}[htb] 
\centering
\includegraphics[width=0.8\textwidth]{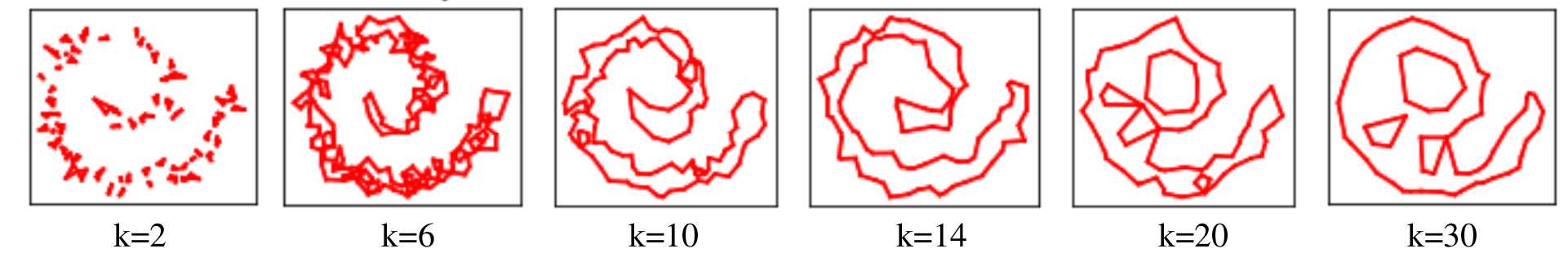}
\caption{The dynamics of envelopes of the spiral manifold with respect to the number of neighbors $k$.} 
\label{fig:neighbor}
\end{figure*}

\textbf{Computation time.} In the feature bagging part, since each base Manifoldron works on separate sub-feature sets, the process-based parallelism can be applied to the Manifoldron for acceleration. Specifically, we run six parallel processes on the CPU with 12 cores. Thus, the training and test costs of the Manifoldron decrease to the $\frac{1}{6}$ as that of the original model. The process number should not exceed the number of cores, as extra processes will overwhelm the CPU and hence adversely slow the model. All the results from these processes are combined to make the final prediction. The training and test time of the Manifoldron on large datasets is listed in Table \ref{tab:time}. The training of all datasets is finished in an acceptable time, which verifies the scalability of the Manifoldron. The test time is measured per sample. Aided by the KD tree, the test time of the Manifoldron is short, which agrees with the theoretical complexity of querying KD tree. Note that the computational cost can be further improved if the Manifoldron runs with more processes on a CPU with more cores. We conclude that the Manifoldron can scale to large datasets.

\textbf{Ablation study.}
In the proposed Manifoldron, the key hyperparameters are the number of neighbors ($k$) in triangulation trimming, the number of base predictors ($N_e$) in feature bagging, and the number of selected features ($N_f$) in feature bagging. Here, ablation studies are done to empirically investigate their influence on model performance.

\underline{Neighbor number}. As shown in Figure \ref{fig:neighbor}, when $k$ is too small, fewer simplices from Delaunay triangulation are retained. This may lead to many small separate envelopes. In the context of the extremely small $k$, no simplices are left because none of them are interior to one another's prescribed neighborhood, and too many edges are trimmed to get simplices. On the contrary, if $k$ is too large, some unnecessary simplices survive, occupying the extra space and hiding the intrinsic manifold structure. Here, we visualize the spiral manifold to show how $k$ changes the shape of envelopes and testify the correctness of our analysis. As Figure \ref{fig:neighbor} shows, the dynamics of the envelopes with respect to the increase of $k$ agrees with our analysis.

\begin{figure}[htb] 
\centering
\includegraphics[width=\linewidth]{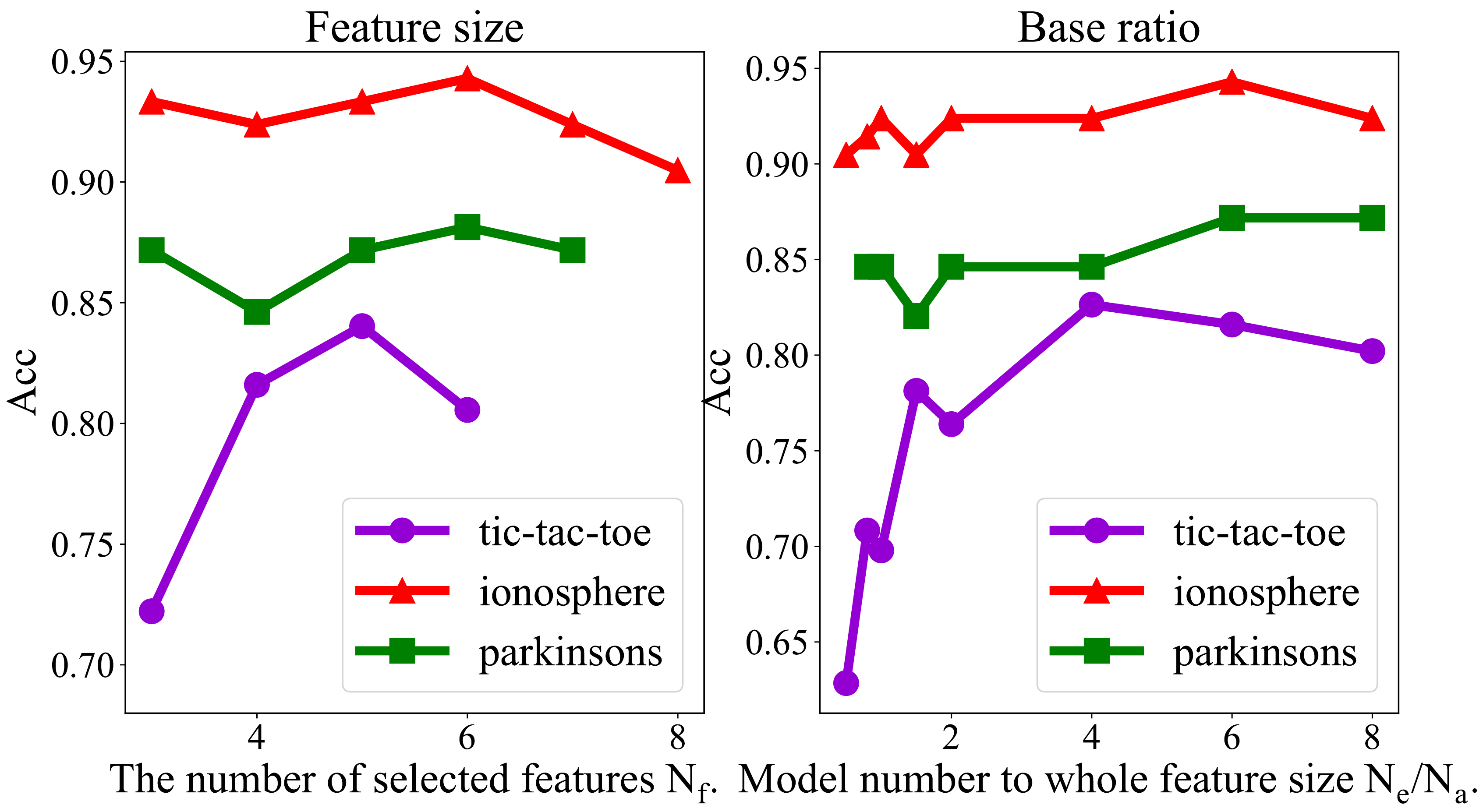}
\caption{Ablation studies on \textit{tic-tac-toe}, \textit{ionosphere} and \textit{parkinsons} with respect to the number of selected features and the number of base predictors to the original feature size.} 
\label{fig:ablation}
\end{figure}

\underline{Feature bagging.}
Feature bagging is employed in the Manifoldron to reduce the computational cost and boost the model performance. In feature bagging, $N_e$ and $N_f$ exert major impact on the computational cost and the performance of the meta Manifoldron. When $N_e$ is small, the entire feature space may not be sufficiently sampled. In contrast, a large $N_e$ may not only introduce superfluous overlaps into the meta Manifoldron but also suffer a high computational cost. Regarding $N_f$, when it is small compared with the total number of features, despite the decreased computational cost of each base Manifoldron, the sampled sub-features may not be representative enough to qualify the task. Consequently, the base Manifoldron trained on them is less desirable, further harming the overall performance of the meta Manifoldron. Meanwhile, although the data with a large $N_f$ is more representative of the original data, a large $N_f$ cannot sufficiently alleviate the curse of dimensionality. Therefore, we need to find an appropriate $N_e$ and $N_f$ to balance the model performance and the computational complexity. Additionally, the selection of $N_e$ and $N_f$ should also maximize the possibility that every sample is sampled. Suppose that the number of all features is $N_{a}$, the possibility that each feature is at least sampled once by the meta Manifoldron is the following:
\begin{equation}
    P = 1 - \bigg(\frac{\binom{N_{a}-1}{N_f}}{\binom{N_{a}}{N_f}} \bigg)^{N_e} = 
    1- \bigg(\frac{N_{a}-N_f}{N_{a}} \bigg)^{N_e}.
\label{poss}
\end{equation}

To evaluate our analysis on $N_e$ and $N_f$, we conduct the ablation studies on three representative datasets: \textit{tic-tac-toe}, \textit{ionosphere}, and \textit{parkinsons}. Figure \ref{fig:ablation} shows the impact of $N_f$ and $N_e / N_a$ on the performance of the Manifoldron on these datasets. For \textit{ionosphere}, the parameters in the given ranges do not cause much fluctuations in performance. For \textit{parkinsons}, the accuracy fluctuates when $N_f$ and $N_e / N_a$ changes. The Manifoldron performs the best when $N_f$ is around 5. Also, The accuracy curves up when $N_e / N_a$ increases and then gradually decreases when $N_e / N_a$ reaches 4.

\section{Real-world Applications}
It was reported that more than 250 different foodborne diseases affect approximately one-third of the world's population annually \cite{stein2017routes}. Certain viable pathogens, growing on food that is exposed to air for a long time, are among the key contributor to transmitting foodborne diseases to humans \cite{stein2017routes}. In this context, the fast and accurate identification of pathogens is a mission-critical problem for public health. Usually, we use a paper chromogenic array (PCA) with a combination of 22 chromogenic dyes \cite{yang2021machine} to probe volatile organic compounds (VOCs) emitted by microorganisms to identify pathogens. The color change of the dye spots profiles the VOC, further indicating the types of pathogens. 


\begin{figure*}[htb] 
\centering
\includegraphics[width=0.8\linewidth]{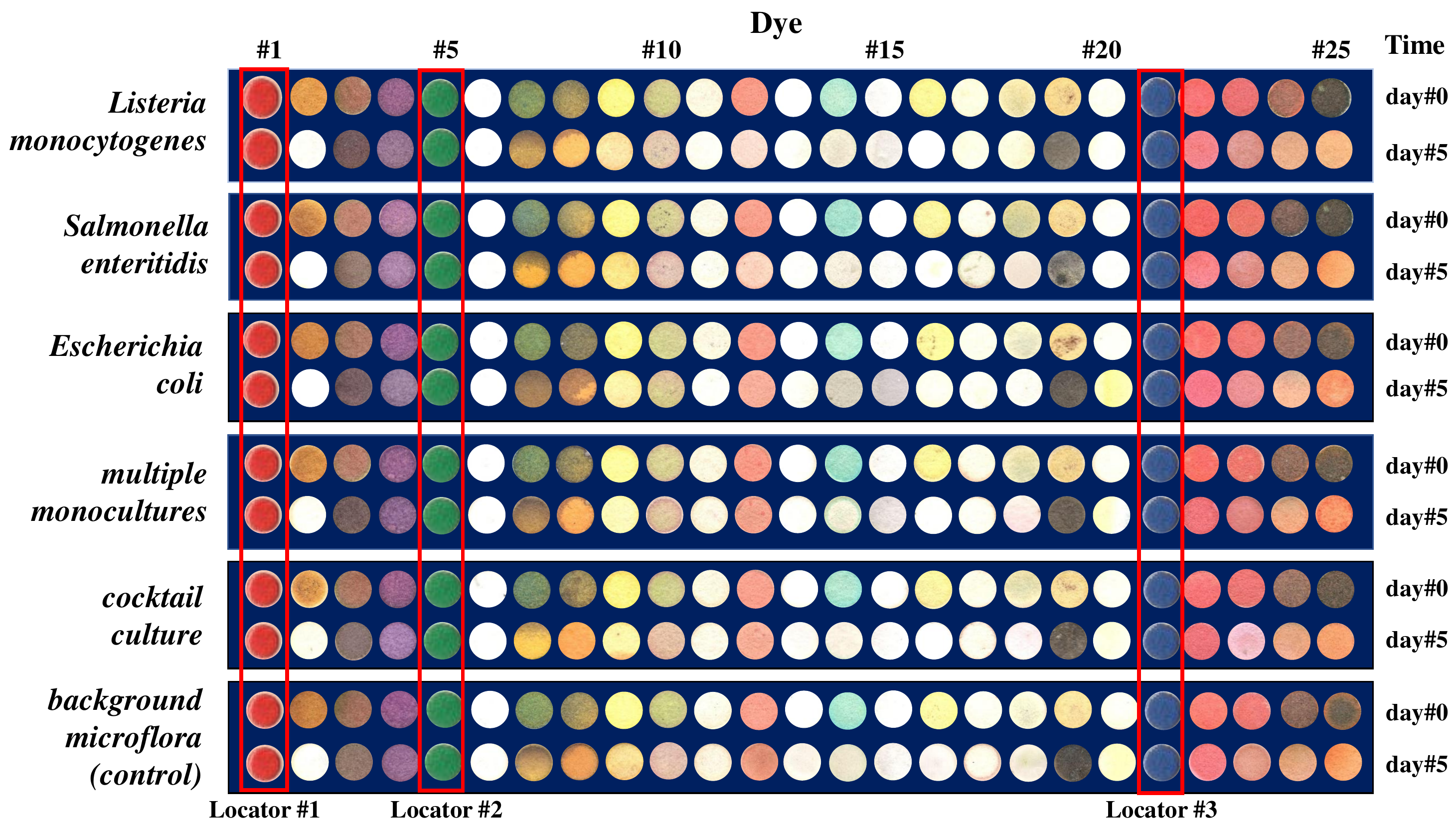}
\caption{The dye color change after exposure to the VOCs informs the type of pathogens. The \textit{background microflora} serves as a control group to probe color change.} 
\label{RGB}
\end{figure*}

Our collaborators have collected a tabular $\triangle \mathrm{R}/\triangle \mathrm{G}/\triangle \mathrm{B}$ dataset by digitizing the color changes in the dyes. Each feature represents an average $\mathrm{R}/\mathrm{G}/\mathrm{B}$ change across all pixels in the dye area. As shown in Figure \ref{RGB}, this dataset has 5 classes that refers to three types of pathogens in single monoculture: \textit{Listeria monocytogenes, Salmonella enteritidis, Escherichia coli}, and two mixture types of pathogens: \textit{multiple monocultures} and \textit{cocktail cultures}. Each class has 36 samples with 66 features that represent the color changes on 22 dyes.


\textbf{Experiments.} 
Since the proposed Manifoldron is a model for the tabular data, we are motivated to apply it to pathogen identification and see if it can solve this real-world problem. The manifoldron is contrasted with the general machine learning models including Nearest Neighbor, Linear-SVM, Decision Tree, Random Forest, Naive Bayes, XGBoost, and  LightGBM, as well as two previously-developed deep learning models: PCA-NN \cite{yang2021machine} and DFFNN \cite{jia2021nondestructive} on pathogen identification. Both PCA-NN and DFFNN were published in prestigious venues in food research.
The PCA-NN and DFFNN are reproduced according to the official codes, while other models follow the same configuration in Section VI. The classification accuracy and macro-F1 score are used to evaluate the models' performances. As shown in Table \ref{application}, compared with its competitors, the proposed Manifoldron delivers the highest accuracy and macro-F1 scores.



\begin{table}[h]
\caption{The results of different models on the pathogen $\triangle \mathrm{R}/\triangle \mathrm{G}/\triangle \mathrm{B}$ dataset. The results are averaged over five runs.}
\centering
\begin{tabular}{|l|c|c|}
    \hline    Method & Accuracy & F1 score \\ 
    \hline
    Nearest Neighbor  & 0.7538 & 0.7635 \\ 
    \hline
    Linear-SVM  & 0.6923 & 0.6985 \\ 
    \hline
    Decision Tree  & 0.2769 & 0.2886 \\ 
    \hline
    Random Forest  & 0.3231 & 0.3328 \\ 
    \hline
    Naive Bayes & 0.5692 & 0.5764 \\ 
    \hline
    XGBoost & 0.6153 & 0.6208 \\ 
    \hline
    LightGBM & 0.6615 & 0.6671 \\ 
    \hline
    PCA-NN & 0.6615 & 0.5888 \\ 
    \hline
    DFFNN & 0.8308 & 0.8278 \\   
    \hline 
    Manifoldron  & \textbf{0.8462} & \textbf{0.8504} \\ 
    \hline
\end{tabular}
\label{application}
\end{table}

\section{Discussion}


\subsection{Comparison with Manifold Regularized Model}
Manifold discovery is an important concept in the Manifoldron. However, the idea of utilizing the manifold can also be realized in other machine learning models as a regularization term \cite{belkin2006manifold}. Aided by manifold regularization, the low-dimensional structures of data are exploited, and the local variance of the model decreases. Generically, the manifold regularization assumes that functions learned from data should be smooth and should not cover the entire input space. Here, we investigate if other machine learning models empowered by manifold regularization can compete with the Manifoldron. 
Due to the popularity of deep learning, we select the neural network (NN) model to compare.

\textbf{Manifold regularization.}
Given a dataset $\{(x_i,y_i)\}_{i=1}^{n}$, a machine learning model $f$ with manifold regularization is derived by solving the following optimization problem:
\begin{equation}
    \argminA_{f \in \mathcal{H}} \frac{1}{n} \sum_{i=1}^n V(y_i,f(x_i)) + \gamma ||f||^2_M,
\end{equation}
where $V(y_i,f(x_i))$ is the empirical loss function, and $\gamma$ is the parameter that controls the magnitude of the regularization. Although there are many choices for $||f||^2_M$, as one of the prominent manifold regularization methods, the Laplacian norm penalizes the smoothness by measuring the total gradients over the manifold \cite{belkin2006manifold}. Mathematically, the Laplacian norm expresses as follows:
\begin{equation}
    ||f||^2_L = \int_{x \in \mathcal{M}} ||\nabla_{\mathcal{M}} f(x)||^2 d \mathcal{P}_X (x),
\end{equation}
where $\nabla_{\mathcal{M}} f(x)$ is the gradient of $f(x)$ over a manifold $\mathcal{M}$, and $\mathcal{P}_X(x)$ is the marginal probability density. Since $\mathcal{P}_X(x)$ is often intractable in real-world tasks, the graph based approach \cite{hein2005graphs} was proposed to approximate the norm: 
\begin{equation}
    ||f||^2_L \approx \frac{1}{(l+u)^2} F^\top L  F,
\end{equation}
where $l$ and $u$ are the labeled/unlabeled sample size, $F = [f(x_1),f(x_2),...,f(x_n)]^\top$, and $L$ is the Laplacian matrix of the graph generated from data. When $u=0$, the task is fully supervised.


\textbf{Experiments.}
In this experiment, we explore the following questions: i) Can the manifold regularization improve the neural network models? ii) Can the neural network with manifold regularization outcompete the Manifoldron? 

We train from scratch a one-hidden-layer fully connected network with 100 neurons. To correctly compute the manifold regularization, we feed all data into the model each time. The model is optimized for 1000 epochs using stochastic gradient descent with a learning rate of $10^{-4}$. The trade-off parameter $\gamma$ is set to 0.01, 0.1, and 1, respectively. We select 7 representative datasets on which the neural network without manifold regularization is inferior to the Manifoldron, thereby we can better answer the abovementioned two questions.

In order to realize manifold regularization, we have to rebuild the NN from scratch instead of using Python sklearn package, which causes the performance of neural networks slightly different from what are shown earlier. Table \ref{tab:manifold_regula} summarizes comparison results. We can see that the regularization only increases the performance on \textit{moons}, \textit{parkinsons}, and \textit{wine}. Furthermore, although manifold regularization boosts the NN performance on \textit{moons}, \textit{parkinsons} and \textit{wine}, it still fails to outperform the Manifoldron. We conclude that manifold regularization does not necessarily contribute to the model performance and can not replace the Manifoldron in terms of leveraging the manifold structure.


\begin{table}[h]
\caption{Performance comparison between NNs with or without manifold regurlaization and the Manifoldron.}
\centering
\scalebox{0.85}{
\begin{tabular}{|l|c|c|c|c|c|}
\hline
    Datasets &  NN & NN ($\gamma$=0.01) & NN ($\gamma$=0.1) & NN ($\gamma$=1) & Manifoldron \\ 
    \hline
    ionosphere & 0.9181 & 0.9029 & 0.9181 & 0.9048 & \textbf{0.9333}\\
    \hline
    magic04 &  0.8162 & 0.8073 & 0.7778 & 0.7998 &  \textbf{0.8183} \\
    \hline
    moons & 0.8644 & 0.8656 & 0.8525 & 0.8678 &  \textbf{0.9625} \\
    \hline 
    parkinson & 0.8093 & 0.8079 & 0.8221 & 0.8051 &  \textbf{0.8813} \\
    \hline 
    satimage &  0.6247 & 0.5805 & 0.5635 & 0.5675 & \textbf{0.7030} \\
    \hline
    tic-tac-toe &  0.7535 & 0.7431 & 0.6944 & 0.7118 & \textbf{0.8298} \\
    \hline
    wine & 0.9074 & 0.9074 & 0.9222 & 0.9259 &  \textbf{0.9629}\\
    \hline 
\end{tabular}}
\label{tab:manifold_regula}
\end{table}

\subsection{Manifoldron as a Regressor}
The Manifoldron can also serve as a regressor. Different from the Manifoldron classifier, the Manifoldron regressor takes all the training data as one manifold for triangulation and trimming, and no decision boundary is needed anymore. The regression for a new point is based on the geometric location of the point relative to a simplex, which is characterized by the barycentric coordinates. 

\textbf{Formulation.} Suppose that the training dataset is $\{(\v_j, y(\v_j)\}_{j=0}^n$, no matter whether a point $\p$ is interior or exterior to a simplex, our heuristic solution is to predict $\p$ with its barycentric coordinates. For one simplex, the prediction is given as
\begin{equation}
    y(\p) = \sum_{i=0}^D \xi_i y(\v_i), \ \ \ s. t. \ \ \ \p = \sum_{i=0}^D \xi_i \v_i.
\label{eqn:basics}    
\end{equation}
Since a simplicial complex is obtained in the Manifoldron, predictions with respect to all simplices should be integrated. Let $\S$ be the simplicial complex, we have
\begin{equation}
    y(\p) = \frac{1}{|\S|} \sum_{\mathrm{S} \in \S} \sum_{i=0}^D \xi_i^{(\mathrm{S})} y(\v_i),
\label{eqn:basic_formula_over_sim}     
\end{equation}
where $|\S|$ is the number of simplices. The above equation is formulated as a sum over all simplices. We can rearrange it as a sum over all vertices, in analogue to the formulation of kernel regression:
\begin{equation}
\begin{aligned}
    y(\p) & = \frac{1}{|\S|} \sum_{j=0}^n \sum_{\mathrm{S} \in \mathcal{S}(\v_j)}  \xi^{(\mathrm{S})}(\p,\v_j) y(\v_j) \\
    & = \frac{1}{|\S|} \sum_{j=0}^n \Big(\sum_{\mathrm{S} \in \mathcal{S}(\v_j)}  \xi^{(\mathrm{S})}(\p,\v_j)\Big) y(\v_j),
\label{eqn:basic_formula}   
\end{aligned}
\end{equation}
where $S(\v_j)$ is a set of simplices consisting of the vertex $\v_j$, and $\xi^{(\mathrm{S})}(\p,\v_j)$ is the barycentric coordinate of $\p$ associated with the vertex $\v_j$ in the simplex $\mathrm{S}$. Recall that an interior point $\p$ inside a triangle splits the triangle into $D+1$ internal simplices. The physical meaning of $\xi_i$ turns out to be the ratio of the area of the internal triangle opposite to $\v_i$ and the area of the original triangle. Such a result can be generalized into exterior points by prescribing that the area of the external triangle is negative. Replacing the barycoordinates in Eq. \eqref{eqn:basic_formula} with the concerning area ratios, we have
\begin{equation}
\begin{aligned}
    y(\p) & = \frac{1}{|\S|} \sum_{j=0}^n \sum_{\mathrm{S} \in \mathcal{S}(v_j)}  \frac{\mathcal{A}^{(\mathrm{S})}(\v_j)}{\mathcal{A}^{(\mathrm{S})}} y(\v_j) \\
    & = \frac{1}{|\S|} \sum_{j=0}^n \Big(\sum_{\mathrm{S} \in \mathcal{S}(v_j)}  \frac{\mathcal{A}^{(\mathrm{S})}(\v_j)}{\mathcal{A}^{(\mathrm{S})}}\Big) y(\v_j),
\label{eqn:basic_formula_area}   
\end{aligned}
\end{equation}
where $\mathcal{A}^{(\mathrm{S})}(\v_j)$ is the area of the opposite triangle of $\v_j$ for the simplex $\mathrm{S}$, and $\mathcal{A}^{(\mathrm{S})}$ is the area of the original triangle \cite{warren1996barycentric}. Eq. \eqref{eqn:basic_formula_area} is a rather basic and elegant regression formula (not another kind of kernel regression), because it cannot be reduced into the typical formula of kernel regression:
\begin{equation}
        y(\p) = \sum_{j=0}^n d(\p, \v_j) y(\v_j),
\end{equation}
where $d(\p, \v_j)$ is some distance measure between $\p$ and $\v_j$.

Furthermore, when a new point and a simplex are far from each other, the impact of the simplex on the point is weak. Therefore, we can discard those distant simplices when predicting $\p$. In the extreme case, we only keep the closest simplex for $\p$ to make prediction. Thus, simplifying Eq. \eqref{eqn:basic_formula_over_sim}, we have
\begin{equation}
y(\p) = \sum_{i=0}^D \xi_i^{(\mathrm{S}_c)} y(\v_i), 
\label{eqn:simplified_formula}    
\end{equation}
where $\mathrm{S}_c$ is the closest simplex to $\p$.

As a feasibility study to verify the performance, we construct datasets based on elementary functions and compare different regressors' fitting errors on these datasets.

\textbf{Data curation.}  We use the following functions to construct datasets:
\begin{equation}
    \begin{cases}
     & f_1 = x^2 + y^2 \\
     & f_2 = \sin(x)+\cos(y) \\
     & f_3 = e^{xy} \\
     & f_4 = x^2 + y^2 + z^2 \\
     & f_5 = e^{xyz} \\
     & f_6 = x^2 + y^2 - z^2.
    \end{cases}
\end{equation}
The input data are generated by a Gaussian distribution with the mean of zero and the standard deviation of one, while the labels are computed with the above functions. We generate 200 samples, and then randomly split them per the ratio of $8:2$ to obtain the training and the test sets, respectively.

\textbf{Compared models.} We compare the proposed Manifoldron with classical regressors such as kernel ridge regressor (KRR), Gaussian process regressor (GPR), k-nearest neighbors regressor (KNNR), decision tree regressor (DTR), MLP regressor (MLP), AdaBoost regressor, and SVM regressor. All the compared models are the mainstream regression models. 

\textbf{Hyperparameters of models.} Different from the Manifoldron as a classifier, here we don't need to deal with the training data class by class. Yet, we take all the training data as one manifold. In the trimming stage, we set the neighbor number of each sample to 14 when querying the KD tree. In the test stage, since this is a simple task, we don't need to use feature bagging to reduce the computational complexity. We take Eq. \eqref{eqn:simplified_formula} to predict, where only the closest simplex is used.  
The experiments are implemented in Python 3.7 on Windows 10 using the Intel i7-8700H processor at 3.2GHz. We set the hyperparameters of other models based on their default configurations. 

\textbf{Results.} 
Table \ref{tab:regressor_quant} summarizes regression results of all models, where the errors for different functions vary considerably because we don't normalize labels in the dataset. As seen, the Manifoldron achieves the lowest errors in fitting all functions, except for $f_4$. Nevertheless, the Manifoldron is only surpassed by the MLP regressor therein. The results of KRR and GPR are always suboptimal. This suggests that it is not desirable to use all training samples to make prediction, as those far samples are relatively irrelevant. We conclude that the Manifoldron can also serve as a good regressor. 

\begin{table}[ht]
\caption{The mean square errors (MSEs) of different models on fitting elementary functions. The best performance on each function is bold-faced.}
\centering
\scalebox{0.75}{
\begin{tabular}{|c|c|c|c|c|c|c|c|c|}
    \hline
    Datasets & KRR & GPR & KNNR & DTR  & MLP & AdaBoost & SVM & Manifoldron\\ 
    \hline
    $f_1$  &   7.820 & 3.376   & 0.595   & 0.238 & 0.372    & 1.025  & 0.327  & \bf{0.101}  \\    
    \hline 
    $f_2$ & 0.508 & 0.275 & 0.012 & 0.043 & 0.017 & 0.052 & 0.024 & \bf{0.006}\\
    \hline 
    $f_3$ & 3.485 & 1.533 & 0.180 & 0.540 & 0.285 & 0.792 & 0.293 & \bf{0.095}\\
    \hline 
    $f_4$ & 17.35 & 7.038 & 2.601 & 1.672 & \bf{0.578} & 2.824 & 1.154 & 0.975 \\
    \hline
    $f_5$ & 47.24 & 43.38 & 37.12 & 45.99 & 31.93 & 44.55 & 42.36 & \bf{18.55}\\
    \hline
    $f_6$ & 8.354 & 6.951 & 2.568 & 8.566 & 0.552 & 2.464 & 2.850 & \bf{0.584} \\
    \hline
\end{tabular}}
\label{tab:regressor_quant}
\end{table}

\subsection{Weaknesses of the Manifoldron for Future Improvements}

Any algorithm has pros and cons. The Manifoldron is no exception. Here, we methodologically analyze the weaknesses of the Manifoldron for future improvements.

\begin{itemize}
\item  \textbf{High computational overhead.} We have come up with feature bagging, KD tree, and point2point projection tricks to reduce the computational cost of the proposed Manifoldron. Now, the Manifoldron can run on the data set with tens of thousands of samples and hundreds of feature dimensions in an acceptable time. However, as the number of samples and the feature dimension go even larger, the Manifoldron tends to be slow. It is still a challenge to scale the Manifoldron to extremely large datasets such as ImageNet. 

\item \textbf{Sensitivity to the number of neighbors.} To discover the true manifold structure, we need to trim the triangles obtained from Delaunay triangulation based on the preset number of neighbors, which is the only hyperparameter of the Manifoldron. The profiles found by the Manifoldron are, however, sensitive to the number of neighbors. In the future, we hope to reduce the sensitivity of the Manifoldron to the hyperparameter and enhance its robustness. 

\item \textbf{Missing value treatment.} In the design of the Manifoldron, the Manifoldron is not endowed with the ability to tackle the missing values. Therefore, the Manifoldron is not applicable to a data set with missing values.

\item \textbf{Outlier sensitivity.} The Manifoldron can be greatly affected by the outliers because Delaunay triangulation and trimming operations avoid isolated vertices. Thus, those outliers still covered by triangles hurt the identification of the true manifold structure. Accordingly, the resultant envelopes will not be sufficiently accurate decision boundaries. Considering that real-world data are often not well-curated, the outlier problem must be resolved in the future.
\end{itemize}

\section{Conclusion}
In this work, we have proposed a novel type of machine learning models referred to as the Manifoldron. Then, to put the proposed Manifoldron in perspective, we have systematically analyzed its characteristics such as manifold characterization capability and its link to a neural network. Lastly, we have compared the Manifoldron with other mainstream machine learning models on 4 synthetic examples, 20 commonly used benchmarks, and 1 real-world biological application. Experimental results have shown that the Manifoldron are better than or comparable to its competitors, and the practical computation time of the Manifoldron is acceptable. In the future, endeavors should be put into translating the Manifoldron into more applications.

\bibliographystyle{ieeetr}
\bibliography{ijcai22.bib}
\end{document}